\documentclass[10pt,twocolumn,letterpaper]{article}

\usepackage[pagenumbers]{cvpr} % To force page numbers, e.g. for an arXiv version

\usepackage{graphicx}
\usepackage{amsmath}
\usepackage{amssymb}
\usepackage{booktabs}
\usepackage{bm}
\usepackage{multirow}
\usepackage{subcaption}
\usepackage{bbding}
\usepackage{booktabs}
\usepackage{gensymb}
\usepackage{balance}
\usepackage[pagebackref,breaklinks,colorlinks]{hyperref}

\usepackage[capitalize]{cleveref}
\crefname{section}{Sec.}{Secs.}
\Crefname{section}{Section}{Sections}
\Crefname{table}{Table}{Tables}
\crefname{table}{Tab.}{Tabs.}

\begin{document}

\title{Joint Video Multi-Frame Interpolation  and Deblurring \\ under Unknown Exposure Time}
\author{Wei Shang$^1$, Dongwei Ren$^1$\thanks{Corresponding author: rendongweihit@gmail.com.} , Yi Yang$^1$, Hongzhi Zhang$^1$, Kede Ma$^2$, Wangmeng Zuo$^{1,3}$\\ 
		$^1$School of Computer Science and Technology, Harbin Institute of Technology\\   $^2$ City University of Hong Kong \quad  $^3$Peng Cheng Laboratory, Shenzhen \\  
}
\maketitle

%%%%%%%%% ABSTRACT
\begin{abstract}
   Natural videos captured by consumer cameras often suffer from low framerate and motion blur due to the combination of dynamic scene complexity, lens and sensor imperfection, and less than ideal exposure setting. As a result, computational methods that jointly perform video frame interpolation and deblurring begin to emerge with the unrealistic assumption that the exposure time is known and fixed.
   In this work, we aim ambitiously for a more realistic and challenging task - joint video multi-frame interpolation and deblurring under unknown exposure time. Toward this goal, we first adopt a variant of supervised contrastive learning to construct an exposure-aware representation from input blurred frames. We then train two U-Nets for intra-motion and inter-motion analysis, respectively, adapting to the learned exposure representation via gain tuning. We finally 
   build our video reconstruction network upon the exposure and motion representation by progressive exposure-adaptive convolution and motion refinement. Extensive experiments on both simulated and real-world datasets show that our optimized method achieves notable performance gains over the state-of-the-art on the joint video  $\times 8$ interpolation and deblurring task. Moreover, on the seemingly implausible $\times 16$ interpolation task, our method outperforms existing methods by more than $1.5$ dB in terms of PSNR.  
   % 	Moreover, our VIDUE not only have better quality in each frame but also are more temporally consistent across frames.  
   %The source code and pre-trained models will be made publicly available. 
\end{abstract}

\begin{figure}[!t]\footnotesize
	\centering
	\setlength{\abovecaptionskip}{3pt} 
	\setlength{\belowcaptionskip}{0pt}
	\begin{tabular}{cccccc}
		\includegraphics[width=0.95\linewidth]{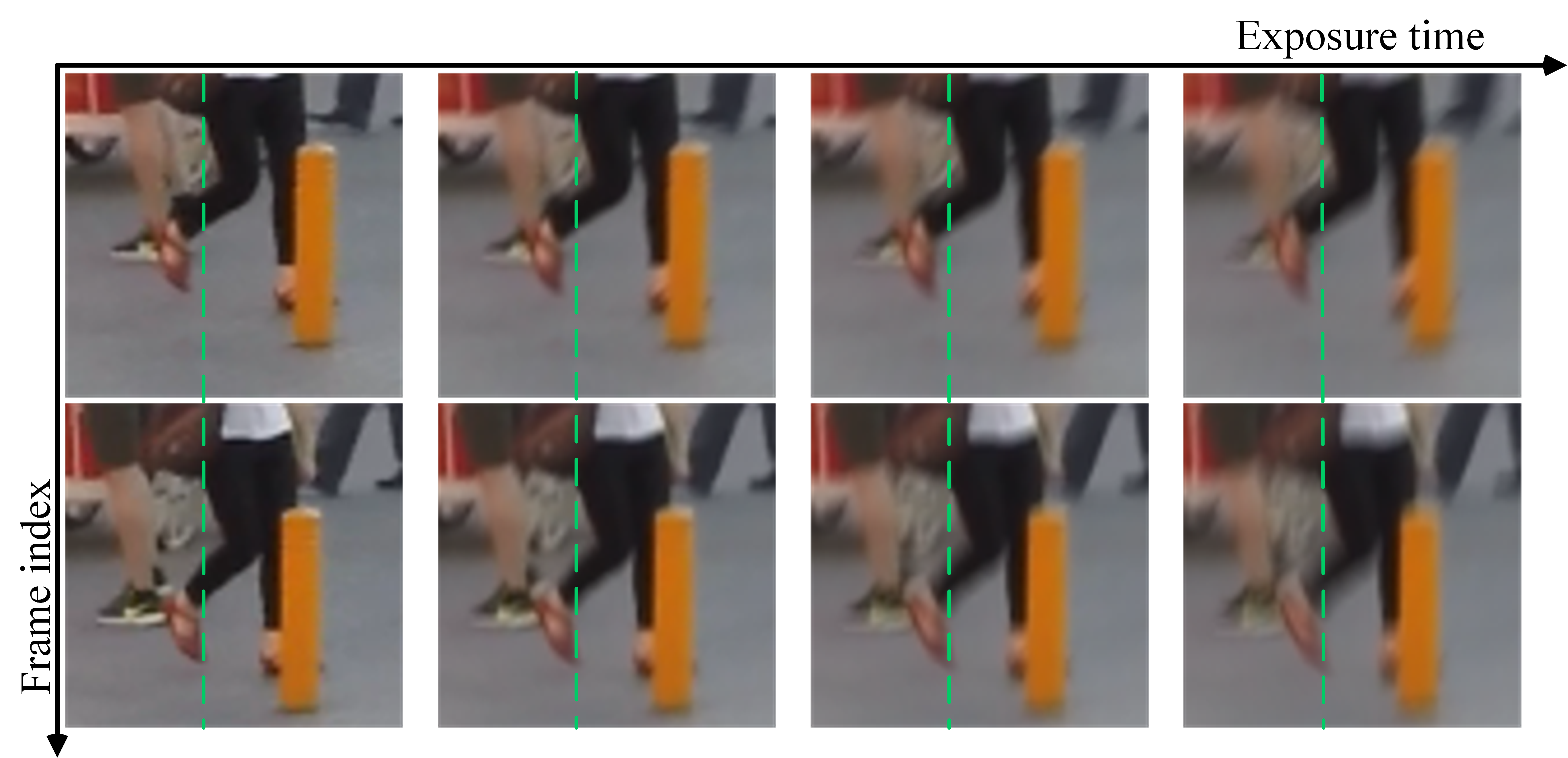}\\
	\end{tabular}
% 	\vspace{-.5em}
	\caption{\small Visualization of motion blur under different exposure time and object motion. Frames from the left to right are obtained with increasing exposure time, leading to severer degree of motion blur. The situation may be worse if stronger motion is present, as illustrated by images in the second row. }
	%Frames from the left to right are captured with increasing exposure time
	\label{fig:exposureill}
	\vspace{-1.5em}
\end{figure}

%%%%%%%%% BODY TEXT
\section{Introduction}
When capturing videos, shutter period (\ie, the inverse of the framerate) and exposure time are two major factors that we are able to manipulate for improved video quality, compared to other confounding factors such as object motion in the scene, lens imperfection, and sensor limitations~\cite{bovik2010handbook}. Particularly, a long shutter period corresponds to lower framerate, and a longer exposure time increases the possibility of introducing severer motion blur (see Fig.~\ref{fig:exposureill}). 
Often, we have good control over the shutter period in the form of the framerate, which is, nevertheless, quite limited in consumer cameras (\eg, $30$ frames per second, or FPS). This is not the case for exposure time, which may constantly and dynamically change depending on the video shooting environment, \eg, illumination and reflection conditions~\cite{kim2021event}. Therefore, video frame interpolation (also known as framerate up-conversion~\cite{castagno1996method,choi2000new,haavisto1989fractional}) and video deblurring methods (under unknown exposure time) are crucial for improving the quality of low framerate blurred videos, and are widely applicable to video editing, video compression, and slow-motion video generation. 
%
% Existing methods try to tackle this issue from two categories: cascade deblurring and interpolation methods and joint deblurring and interpolation methods.
%

In literature, video frame interpolation~\cite{lee2020adacof,kalluri2023flavr,xu2019quadratic,sim2021xvfi} and video deblurring~\cite{pan2020cascaded,cho2021rethinking} have long been treated as individual problems and tackled separately with worth-celebrating successes.
A straightforward approach to joint video frame interpolation and deblurring is to deploy deblurring methods followed by frame interpolation.
However, this type of cascaded methods usually cannot obtain satisfactory reconstruction results, 
% Given a blurry video, deploying deblurring frameworks followed by interpolation methods to predict sharp intermediate frames is not optimal 
since algorithm-dependent deblurring artifacts would be propagated to and amplified in interpolated frames~\cite{shen2020blurry}. Similar situations will occur if we cascade video frame interpolation first~\cite{argaw2021motion}.   
%
%First, the interpolation performance is highly dependent on the quality of the deblurred images. Second, given imperfect deblurred frames in the cascade scheme, the interpolation model with a short temporal scope can hardly maintain the long-term motion consistency among adjacent frames.
%
%
This inspires recent work~\cite{Oh2022DeMFI,shen2020blurry,argaw2021motion,xu2021temporal} to cast video frame interpolation and deblurring (or super-resolution) as a joint and emerging low-level vision problem. 
% to jointly recover sharp frames and temporally upsample frames from blurry sequences , 
%
However, these methods assume known and fixed exposure time in the video degradation model, which is unrealistic, especially when the auto-exposure mode is enabled. Thus, they are bound to generalize
poorly in the real-world.
% yet there are very limited studies handling joint deblurring and interpolation problem under arbitrary exposure time setting.	
%

Currently, there are two studies~\cite{zhang2020video,kim2021event} considering the setting of unknown exposure time. Zhang \etal~\cite{zhang2020video} computed generic quadratic  motion trajectories from consecutive blurred (or estimated sharp) frames.
% for video interpolation and deblurring.
%
They directly cascaded existing deblurring and frame interpolation methods, suffering from the error propagation problem. 
% me suffers sub-optimal solution due to artifacts propagation between deblurring and interpolation.
%
Benefiting from additional event cameras~\cite{brandli2014240}, Kim~\cite{kim2021event}
proposed an event-guided and end-to-end optimized video deblurring method under unknown exposure time, but event sensors have not been equipped on consumer cameras, restricting its practical applications.
% .  but event cameras are not so popular and the method cannot handle xxx. 
% .But event data is difficult to obtain due to the expensive event sensor.

%
In this work, we aim ambitiously for the more realistic and challenging task - joint video multi-frame interpolation and deblurring under unknown exposure time. Our primary design philosophy is \textit{adaptive computation}: we adapt our Video frame Interpolation and Deblurring method under Unknown Exposure time (VIDUE) to exposure-aware and motion-aware representations, where the computation of the motion-aware representation is further adapted to the exposure-aware representation. To achieve this, we first adopt a variant of supervised contrastive learning to extract an exposure-aware representation from a low framerate blurred video.  We then train two U-Nets: one is responsible for intra-motion analysis (\eg, assessing motion complexity within each frame); the other is responsible for inter-motion analysis (\eg, assessing motion continuity between frames). We adapt the two U-Nets to the exposure-aware representation via gain tuning~\cite{mohan2021adaptive} (also can be seen as a variant of sequence-and-excitation~\cite{hu2018squeeze}). 
% \red{When unknown exposure time is modeled, motion between adjacent frames is more easily captured.} 
We last develop a video reconstruction network with a U-Net-like structure, which enables exposure-adaptive convolution and motion refinement in a progressive fashion.

Extensive experiments on both synthetic and real datasets show that the proposed VIDUE consistently produces higher-quality deblurred and $\times 8$ interpolated frames both visually and in terms of standard quality metrics. Moreover, VIDUE exhibits significant performance gains ($\ge 1.5$ dB) over the state-of-the-art on the seemingly implausible $\times 16$ interpolation and deblurring task.

\section{Related Work}
% In this section, we provide a concise overview of video frame interpolation and deblurring methods, with an emphasis on deep learning-based ones. 

\subsection{Video Frame Interpolation}
The video industry has long been interested in increasing the temporal resolution of a video sequence, \ie, framerate up conversion~\cite{chen1998frame,choi2000new,jeon2003coarse}, with primary application to video compression. The problem resurges in the computer vision community under the name of video frame interpolation along with the renaissance of deep learning. Both Center-Frame Interpolation (CFI)~\cite{niklaus2017video,choi2020channel,bao2019depth,lee2020adacof} and Multi-Frame Interpolation (MFI)~\cite{xu2019quadratic,huang2022rife,jiang2018super} have been extensively investigated. For 
center-frame interpolation, Niklaus \etal~\cite{niklaus2017video} proposed separable kernel prediction networks to handle large motion, optimized by ``perceptual'' losses~\cite{johnson2016perceptual}.
Bao \etal~\cite{bao2019depth} proposed to incorporate depth information during interpolation to combat occlusion through bidirectional flow estimation.
Lee~\etal~\cite{lee2020adacof} combined and generalized the kernel-based and flow-based methods by offset prediction, and introduced an adversarial loss to examine the naturalness of the interpolated frame w.r.t. adjacent input frames. For multi-frame interpolation, more complex motion trajectory models need to be specified compared to the linear model assumption typically used in center-frame interpolation.
Xu~\etal~\cite{xu2019quadratic} proposed a quadratic interpolation scheme to allow the inter-motion to be curvilinear. 
Chi~\etal~\cite{chi2020all} proposed a cubic-based motion model with a relaxed warping loss to further boost interpolation accuracy for complex motion scenes.
Huang~\etal~\cite{huang2022rife} designed the so-called privileged distillation scheme for real-time arbitrary timestamp frame interpolation.
In addition, Kalluri~\etal~\cite{kalluri2023flavr} cast multi-frame interpolation as a self-supervised pretext task to benefit downstream video applications, such as action recognition and video object tracking. 3D convolutions were adopted for spatiotemporal feature extraction.
All these methods would encounter difficulties when processing motion-blurred videos because the optical flow/motion estimation as the core module will become less accurate.

\subsection{Image and Video Deblurring}
%\vspace{-0.6em}
Traditionally, image deblurring is formulated as a Maximum A Posteriori (MAP) problem, which relies heavily on natural image priors, such as total variation~\cite{chan1998total}, smoothness priors based on Markov random fields~\cite{raj2005graph}, normalized sparsity~\cite{krishnan2011blind}, and  color-line priors~\cite{lai2015blur}. Recent image deblurring methods adopt a pure data-driven approach, learning to deblur from massive deblurred-clean image pairs~\cite{zhang2019deep, cho2021rethinking}. Among popular architectural design choices, coarse-to-fine estimation has been extensively studied~\cite{gao2019dynamic, Nah_2017_CVPR, park2020multi, cho2021rethinking}. For example, Chi~\etal~\cite{cho2021rethinking} adopted a U-Net to accept blurred images of multiple scales, and produce the corresponding set of sharp images in parallel. Alternatively, Ren~\etal~\cite{ren2020neural} proposed an unsupervised image deblurring scheme by leveraging deep priors~\cite{ulyanov2018deep} of both underlying sharp images and blur kernels.
%
%Cho et al.~\cite{cho2021rethinking} presented a single UNet that has distinct features, enabling much simpler but more effective coarse-to-fine deblurring.
%
For video deblurring, the added temporal dimension can be coped with recurrent computation~\cite{hyun2017online,zhong2020efficient}, optical flow estimation~\cite{kim2018spatio,pan2020cascaded,shang2021bringing}, and deformable convolution~\cite{wang2019edvr}, aided by handcrafted priors~\cite{pan2020cascaded} or complementary modalities~\cite{shang2021bringing}.
%Kim et al.~\cite{hyun2017online} develop a spatial-temporal recurrent network with a dynamic temporal blending layer for latent frame restoration.
%
%To better leverage spatial and temporal information, Kim et al.\cite{kim2018spatio} introduced an optical flow estimation step for aligning and aggregating information across the neighboring frames to restore latent clean frame. 
%
%In~\cite{wang2019edvr}, Wang et al. developed deformable convolution in pyramid manner to implicitly align adjacent frames for better leveraging temporal information.   
%
%Recently, Pan et al.~\cite{pan2020cascaded} proposed to simultaneously estimate the optical flow and latent frames for video deblurring with the help of temporal sharpness prior. 
%
%Zhong et al.~\cite{zhong2020efficient} adopted residual dense blocks into RNN cells, so as to efficiently extract the spatial features of the current frame.
%
%Shang et al.~\cite{shang2021bringing} found that some frames in a video with motion blur are sharp, and proposed to detect sharp frames in video and then restore the current frame by the guidence of sharp frames. 
%
Image/video deblurring can only restore existing blurred frames, even  motion information between input frames are computed to facilitate deblurring, which is kind of computational waste.
Therefore, it is rational to consider joint video frame interpolation and deblurring, especially when the motion between consecutive frames is computed in either case.

\subsection{Joint Video Interpolation and Deblurring}
%\vspace{-0.6em}
Recently, some researchers began to address the problem of joint video frame interpolation and deblurring. 
Shen~\etal~\cite{shen2020blurry} proposed a pyramidal method for center-frame interpolation and deblurring. %
Argaw~\etal~\cite{argaw2021motion} adopted a motion-based approach for not only multi-frame interpolation but also extrapolation.
Oh~\etal~\cite{Oh2022DeMFI} introduced flow-guided attention and recursive feature refinement to improve the reconstruction performance.
All these methods assume known and fixed exposure time, which is less realistic. Closest to ours, Zhang~\etal~\cite{zhang2020video} chose to cascade deblurring and interpolation networks under unknown exposure time. However, the performance would inevitably be compromised by the inaccuracy of the optical flow module in the deblurring network.
%To handle this problem effectively, five works [13, 19, 40, 41, 61] delicately have shown that
%joint approach is much better than the cascade of two separate tasks such as deblurring and VFI, which may lead to sub-optimal solutions.
Here, the proposed VIDUE is designed to adapt its computation to exposure-aware and motion-aware representations for joint video multi-frame interpolation and deblurring under unknown exposure time. 
%We propose a temporal feature extractor for implicitly learning the discriminative temporal condition.
%Then we embed the temporal condition into our restoration network as a condition. 

\section{Proposed VIDUE}\label{sec:degrad}
In this section, we first present the problem formulation of joint video frame interpolation and deblurring under unknown exposure time, and then describe in detail the proposed VIDUE method, consisting of an exposure-aware feature extractor $g_e$, an intra- and inter-motion analyzer $g_a$, and a video reconstruction network $f$.

\subsection{Problem Formulation}
%In this section, we introduce the degradation model for low framerate motion blur, and we formulate the video deblurring and interpolation with arbitrary exposure and readout.

%\subsection{Degradation Model}
When capturing a video frame, the shutter period $\Delta t$ includes two phases: the exposure phase $\Delta t_e$ and the effective readout phase $\Delta t_r$, where $\Delta t = \Delta t_e + \Delta t_r$.
Given a video with $T$ frames $\bm{y} = \{\bm{y}_t\}_{t=1}^T$, the $t$-th frame $\bm{y}_t$ is essentially an integral of the latent image $\bm{x}_\tau$ at each instant time $\tau$ over the exposure time $\Delta t_e$:
%
%
%acquising A video frame acquisition consists of the exposure phase $\Delta t_e$ and effective readout phase $\Delta t_s$. 
%The summation of two phases (shutter period), represents the time to acquire one video frame ($\Delta t = \Delta t_e + \Delta t_s$). 
%By the nature of frame-based cameras, motion blur only occurs during the exposure phase. 
%A blurry video with $T$ frames $\bm{Y} = \{\bm{y}_t\}_{t=1}^T$.
%%
%Formally, we assume that there exists a latent image $\bm{x}_{\tau}$ at each instant time $\tau$.
%A blurry frame $\bm{B}_t$ is generated by integrating the latent images over the exposure interval $\Delta t_e$, 
%% to obtain one captured frame. The discrete degradation model for blurriness is generally formulated as follows:
\begin{equation}\label{eq:gen}
	\bm{y}_t=\frac{1}{\Delta t_e} {\int}_{\tau=t
 \cdot\Delta t}^{\tau=t\cdot\Delta t + \Delta t_e } \bm{x}_{\tau} d{\tau}. \\
\end{equation}
%$\bm{B}_t$ is the $t$-th acqui/red low-frame-rate blurry frame, and $T$ is the video length. 
%
For consumer-grade cameras, the captured videos for dynamic scenes usually suffer from low framerate and blur due to long shutter period with a large portion of exposure time. 
Joint video $\times S$ interpolation and deblurring aims to recover a high framerate video with sharp frames: $\bm{x} = \{\bm{x}_t\}_{t=1}^{T\times S}$  from the observed video $\bm{y}$. In this work, we assume the shutter period $\Delta t$ is known, while the exposure time $\Delta t_e$ is not, and thus $\Delta t_e$ is directly linked to the strength of the motion blur.
It is noteworthy that the formulation in Eq.~\eqref{eq:gen} is different from DeMFI~\cite{Oh2022DeMFI}, which synthesizes blur by directly averaging consecutive frames, while ignoring the readout phase.
% In existing methods, video deblurring and video interpolation have been individually studied, while joint video deblurring and interpolation is rarely studied. 
%
% In most existing methods, readout phase is ignored, i.e., $\Delta t_e = \Delta t$ is simply assumed. 
% %
% In this regard, latent images at readout phase are actually not considered, and the reconstructed videos may suffer from temporal inconsistency. 
\subsection{Exposure-Aware Feature Extractor}
Our first design choice is that the video reconstruction network $f$ should adapt to different exposure time. For the $\times S$ interpolation task, we have the same number of $S$ exposure time durations, where $\Delta t_e \in\{1,2,\ldots, S\}$ and $\Delta t = S$ in Eq.~\eqref{eq:gen}. $\Delta t_e : \Delta t = S:S$ means that the effective readout time $\Delta t_r$ is zero. That is, the actual readout phase is fully overlapped with the next exposure phase. 

We work with a mini-batch of input video sequences, from which we create two multiviewed versions by applying horizontal and vertical flipping, 90{\degree} rotation, and random cropping  to obtain $\mathcal{B} =\{(\bm{y}^{(i)}, \Delta t_e^{(i)})\}_{i=1}^{\vert\mathcal{B}\vert}$ with $\bm{y}^{(i)}\in \mathbb{R}^{(T\times 3) \times H\times W}$. We combine the temporal and channel dimensions into one to na\"{i}vely enable spatiotemporal analysis using 2D convolutions. We adopt a variant of ResNet18 as $g_e: \mathbb{R}^{(T\times 3)\times H \times W} \mapsto \mathbb{R}^{C\times 1}$, where we replace the classification head with two Fully-Connected (FC) layers with leaky ReLU nonlinearity in between, to extract the exposure-aware feature representation $\bm{u}^{(i)}\in \mathbb{R}^{C\times 1}$.

%
% For temporal upsampling factor $S$, there are $S$ exposure settings, i.e., ${v}$ can be $\alpha:S$, where exposure time $\alpha$ can be set as 1,2,$\cdots$,$S$. And $S$:$S$ means readout phase is totally overlapped with next exposure phase. We set $\alpha$ as an integer to generate corresponding training data. 
%1:$S$, 2:$S$, $\cdots$, $S$:$S$,
%
%Borrowing from supervised contrastive learning\cite{khosla2020supervised}, we utilize supervised contrastive learning to learn the distinguishing features under different exposure time settings. 
%
%For each input sequence $\mathbf{B}$, we generate two random augmentations, $\tilde{\mathbf{B}} = \emph{Aug}(\mathbf{B})$, each of which represents a different view of the sequence and contains some subset of the information in the original sequence.
%

% \begin{equation}
% 	\bm{v} = \mathcal{T}({\bm{Y}}),\quad
% 	\bm{z} = \mathcal{P}(\bm{v})
% \end{equation}
% where $\mathcal{T}$ is a simple encoder to map input frames to 
%

\begin{figure*}[!t]\footnotesize
	\centering
	\setlength{\abovecaptionskip}{3pt} 
	\setlength{\belowcaptionskip}{0pt}
	\begin{tabular}{cccccc}
		\includegraphics[width=0.9\linewidth]{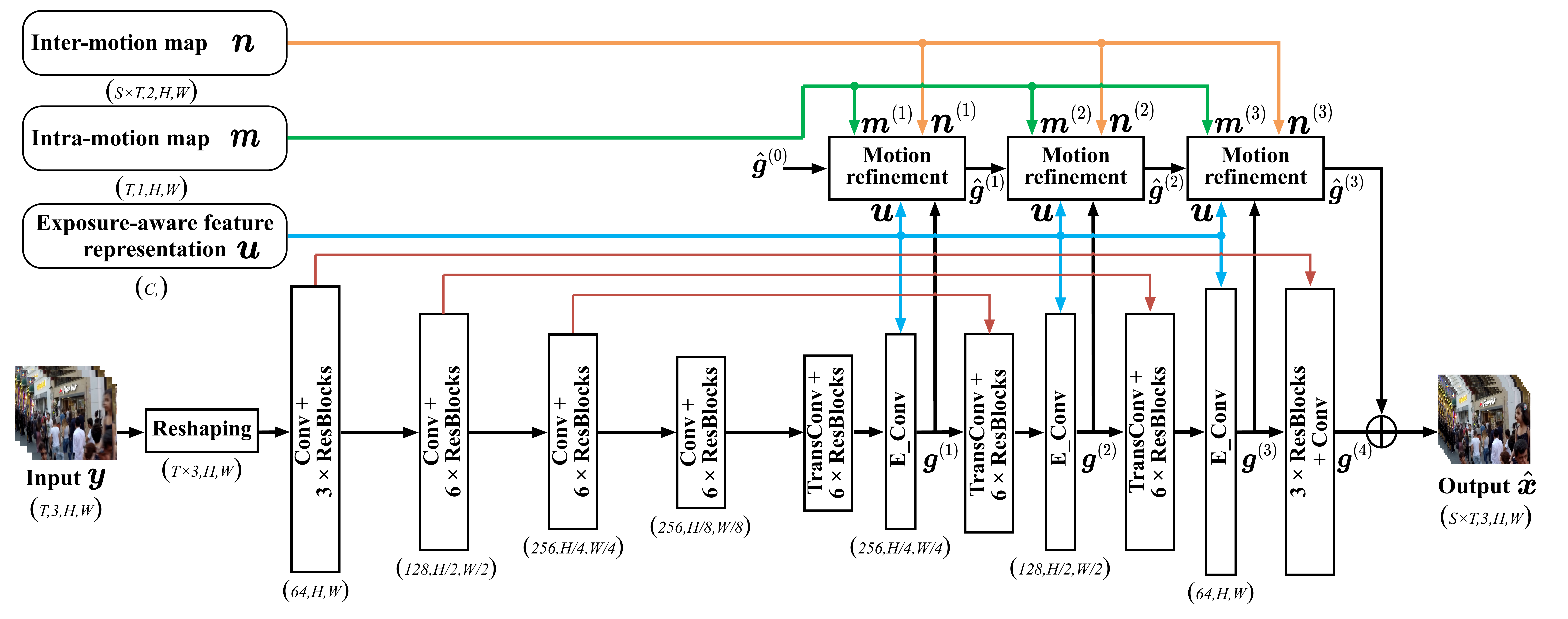}\\
	\end{tabular}
	\vspace{-1em}
	\caption{Overview of the reconstruction network of VIDUE, by which low framerate blurred video $\bm y$ with $T$ frames can be reconstructed to high framerate sharp video with $S\times T$ frames for the $\times S$ interpolation and deblurring task. The reconstruction network is built upon the exposure-aware representation $\bm{u}$ and motion-aware representations $\bm{m}$ and $\bm{n}$ by exposure-adaptive convolution, and we implement progressive motion refinement for better reconstruction performance.
	}
	\label{fig:framework}
\end{figure*}
We want to make the feature representations $\{\bm u^{(i)}\}$ corresponding to different exposure time to be as discriminative as possible. Thus we resort to supervised contrastive learning~\cite{khosla2020supervised}, and introduce the relative weighting to indicate the difference in exposure time between each sample and the anchor:
\begin{equation}\small\label{eq:contr}
	\ell_{\mathrm{ws}}=\sum_{\bm u \in \mathcal{B}} \frac{-1}{|\mathcal{P}|} \sum_{\bm{v} \in \mathcal{P}} \log \frac{\exp \left(\bm{u}^\intercal \bm{v}/ \alpha\right)}{\sum\limits_{\bm{v}' \in \mathcal{B}\setminus\{\bm{u}\}} {w}(\bm{u},\bm{v}') \cdot \exp \left(\bm{u}^\intercal \bm{v}' / \alpha\right)},
\end{equation}
where $\bm{u}$ denotes the anchor, $\mathcal{P}$ contains positive samples that share the same {exposure} time with the anchor, and $\alpha$ is a fixed temperature parameter. The relative weighting can be straightforwardly computed by $w(\bm{u}^{(i)}, \bm {u}^{(j)}) = \left\vert \Delta t_e^{(i)} - \Delta t_e^{(j)}\right\vert$.
%
% By adopting the weighted contrastive loss, 
% more different exposure time are forced to have more discriminative features. 
%
% the feature representations $\{\bm{u}^{(i)}\}$ are expected to be as disctiminative as possible for different exposure time. 
%
We also try to formulate the exposure-aware feature representation learning as ordinal regression~\cite{gutierrez2015ordinal}, 
%
% \red{(but obtain significantly worse final results, which may be attributed to the neural collapse phenomenon~\cite{papyan2020prevalence})$\rightarrow$(
but obtain worse final results (see Table~\ref{tabel:tp}). 
% The reason may be attributed to the extra training difficulty brought by exactly predicting exposure time in ordinal regression.
% ). }
%
After training, the exposure-aware feature extractor $g_e$ is fixed during the training of the motion analyzer and the final reconstruction network. 
%Our goal is utilizing a unified architecture to enhance the input video under arbitrary exposure time. Given a blurry sequence of length $N$, we can restore $\Delta t$ sharp frames corresponding to the center timestamp of input blurry sequence.
%
%In the following, we use $\mathbf{B}$ to depict the input blurry sequence, use $\hat{\mathbf{I}}$(or $\mathbf{I}$) to depict the restored sharp sequence(or the corresponding ground-truth sequence) for simplicity. 
%
%We formulate the deblurring and interpolation with arbitrary exposure time problem as maximizing a posteriori of the output frames conditioned on the blurry frames and exposure time setting:
%\vspace{-0.4em}

%where ${v}$ denotes the exposure setting of captured video $\bm Y$. %optimal joint space-time enhancement model. We propose to use trainable neural networks to approximate the optimal model $\mathcal{M}^{\star}$.
\begin{figure}[!t]\footnotesize
	\centering
	\setlength{\abovecaptionskip}{3pt} 
	\setlength{\belowcaptionskip}{0pt}
	\begin{tabular}{cccccc}
		\includegraphics[width=0.7\linewidth]{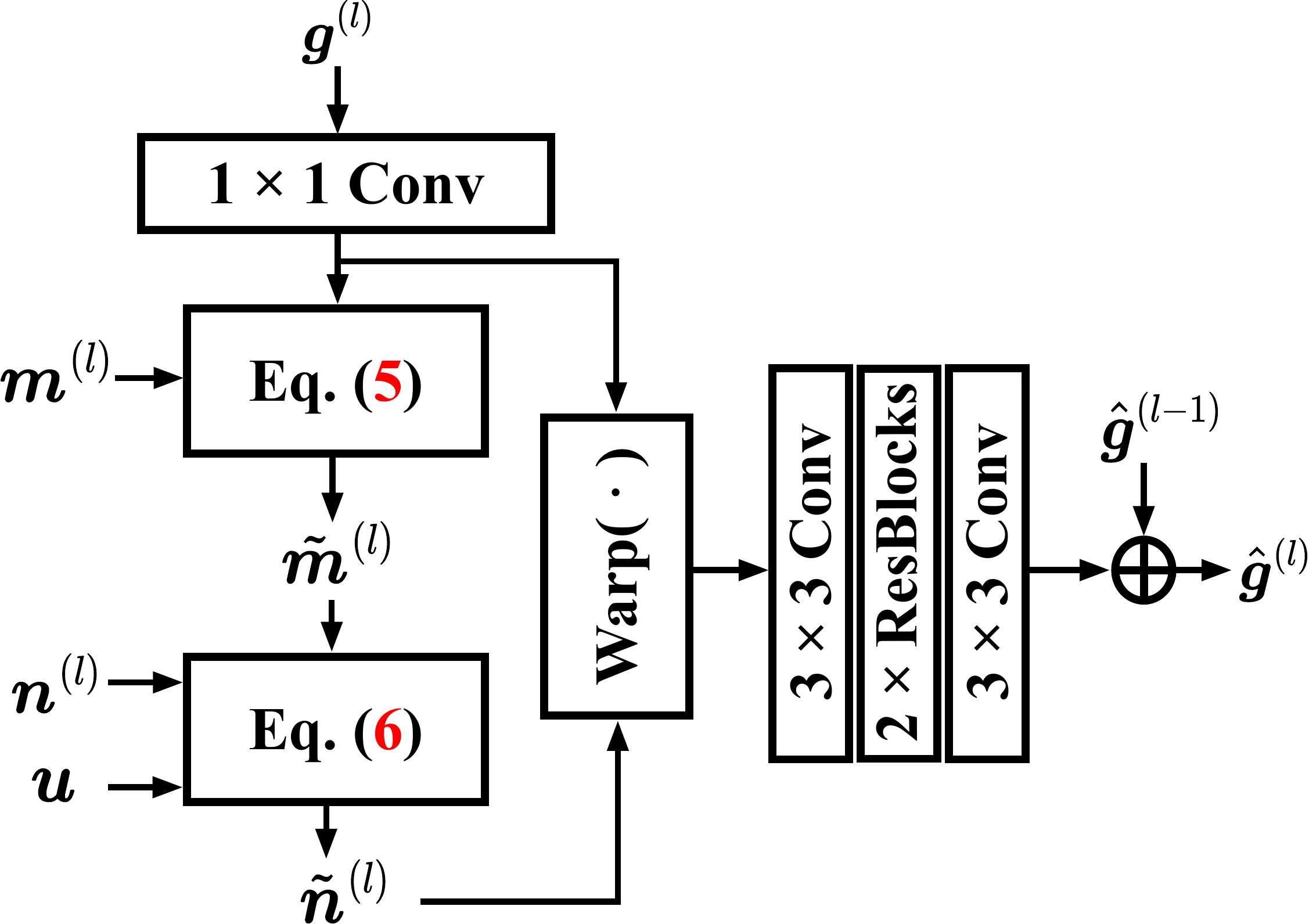}\\
	\end{tabular}
% 	\vspace{-1em}
	\caption{Overview of the proposed progressive motion refinement module. We first refine the intra-motion maps $\bm{m}^{(l)}$ with the help of $\bm g^{(l)}$ to obtain $\tilde{\bm{m}}^{(l)}$, and then refine the inter-motion maps $\bm{n}^{(l)}$ based on  $\tilde{\bm{m}}^{(l)}$ and  $\bm{u}$. Finally, we obtain the motion-refined output feature $\hat{\bm{g}}^{(l)}$ by warping the input feature $\bm{g}^{(l)}$  according to $\tilde{\bm{n}}^{(l)}$, followed by further refinement and incorporation of the upsampled $\hat{\bm{g}}^{(l-1)}$ from the previous stage via addition.
	}
	\label{fig:g_a}
\end{figure}

\subsubsection{Intra- and Inter-Motion Analyzer}
%Our motion estimation $\mathcal{F}_1$ includes motion map estimation and inter and intra optical flows estimation.
Our second design choice is that the video reconstruction network $f$ should adapt to different motion patterns presented in dynamic scenes. We choose to analyze both intra-motion within each video frame, which is relevant to motion complexity (captured in a given exposure time period $\Delta t_e$), and inter-motion between frames, which is pertinent to motion continuity (captured in a given shutter period $\Delta t$). Our motion analyzer $g_a: \mathbb{R}^{(T\times 3)\times H\times W} \mapsto \mathbb{R}^{T\times 1\times H\times W}\times \mathbb{R}^{(S\times T)\times2\times H\times W}$, computes, from an input video sequence $\bm{y}^{(i)}$, $T$ intra-motion maps and $S\times T$ inter-motion maps of the same spatial size, respectively, whose computation is adaptive to the exposure-aware representation $\bm{u}^{(i)}$.

\noindent\textbf{Intra-Motion Analysis}. We adopt a pre-trained light-weight U-Net~\cite{zhang2021exposure}, in which we tune the ``gain'' (\ie, a single multiplicative scaling parameter) of each channel of the intermediate representations $\bm{z}\in\mathbb{R}^{C'\times H'\times W'}$ (in the expansive path~\cite{ronneberger2015u}):
\begin{align}\label{eq:gaint}
% \begin{aligned}
\bm g_{i\bullet} = \bm{u}'_i \cdot \bm{z}_{i\bullet}, \quad \bm{u}' = \sigma\left(\mathbf{W}_2^\intercal \mathrm{LReLU}\left(\mathbf{W}_1^\intercal \bm{u}\right)\right).
% \end{aligned}
\end{align}
Here $\bm{z}_{i\bullet}$ stands for the $i$-th channel, and $\bm{u}'\in\mathbb{R}^{C'\times 1}$ is the gain vector computed from the exposure-aware representation $\bm{u}$ by two FC layers with leaky ReLU in between, followed by a Sigmoid function $\sigma(\cdot)$. $\{\mathbf{W}_1, \mathbf{W}_2\}$ are learnable weight matrices. Eq.~\eqref{eq:gaint} can also be seen as a form of the squeeze-and-excitation operation~\cite{hu2018squeeze}, where we ``squeeze'' the raw video sequence into $\bm u'$ through $\bm u$, and use it to ``excite'' $\bm{z}$.
The results from the U-Net are intra-motion offsets $\bm{o}^{(s)}\in\mathbb{R}^{T\times 2 \times H\times W}$ and $\bm{o}^{(e)}\in \mathbb{R}^{T\times 2 \times H\times W}$, which are the starting and ending positions of estimated motion trajectories in  horizontal and vertical directions, respectively. We compute the final intra-motion maps $\bm{m}\in\mathbb{R}^{T\times 1\times H\times W}$ by subtracting $\bm{o}^{(e)}$ from $\bm{o}^{(s)}$, followed by root mean squared of the subtracted offsets.
Intra-motion maps can also be integrated into $g_e$ to improve representation capabilities.
% an existing motion map estimation method ETR~\cite{zhang2021exposure}, which is a
% We use the pre-trained model provdied by~\cite{zhang2021exposure}, which is trained on GoPro dataset.
%where the estimated motion map provides the blur degree in spatial dimension to guide the video restoration. As demonstrated in ETR\cite{zhang2021exposure}, the motion offset estimation can be efficiently solved by a neural network,
% \begin{equation}
	% 	\bm{o}^{s}, \mathbf{o}^{e} = \emph{ETR}(\bm{Y}),\\
	% \end{equation}
% where $\emph{ETR}(\cdot)$ is pre-trained motion offsets estimation network in~\cite{zhang2021exposure}, $\bm{o}^{s} \text{ and } \bm{o}^{e}$ are the shift of pixels at start and end state of exposure phase corresponding to input blurry frames, respectively.
% First, we get motion offsets $\bm{o}^{s} \text{ and } \bm{o}^{e}$ with temporal dimension $T$ by pre-trained ETR, where $\bm{o}^{s} \text{ and } \bm{o}^{e}$ are shift of pixels at start and end of exposure phase of input sequence $\bm{Y}$, respectively.
%

% \begin{equation}
	% 	\mathbf{o} = |\mathbf{o}^{s} - \mathbf{o}^{e}|,\\
	% \end{equation}
% \begin{equation}
	% 	\mathbf{m} = \sqrt{({\Delta \alpha})^2 + ({\Delta \beta})^2},\\
	% \end{equation}
% $\mathbf{m}$ is the spatial offset distance of each pixel during the exposure time of bulrry frames, which can be used to guide the restoration network $\mathcal{F}$ to distinguish the blur degree of input sequence in spatial dimension.
%
%\textbf{Inter-motion Estimation.}

\begin{table*}[!htb]\footnotesize  %[t]  %
	\centering
	\setlength{\abovecaptionskip}{0pt} 
	\setlength{\belowcaptionskip}{0pt}
	\begin{tabular}{l|c|c|c|c|c|c|c}
		\hline
		
		\hline	
		\multirow{2}{*}{Method} & {Inference time (sec)}  &\multicolumn{3}{c|}{GoPro-5:8}  &  \multicolumn{3}{c}{GoPro-7:8} \\
		\cline{3-8}
		& / Framerate (FPS) & Deblurring  & Interpolation & Avg & Deblurring  & Interpolation & Avg\\ %Total\_
		\hline
		CDVD-TSP+QVI & 1.02 / 7.84  &\multirow{2}{*}{33.06 / 0.952} &  30.02 / 0.914  &  30.40 / 0.918   &   \multirow{2}{*}{31.39 / 0.933}   &  29.39 / 0.902  & 29.64 / 0.906   \\
		CDVD-TSP+RIFE & 0.40 / 20.00 &  &  29.95 / 0.907  &  30.34 / 0.913   &     &  29.15 / 0.892
		&  29.44 / 0.897   \\
		
		MIMOUNetPlus+QVI & 0.78 / 10.26 &\multirow{2}{*}{34.42 / 0.961} &  30.48 / 0.920
		&  30.97 / 0.925  &  \multirow{2}{*}{33.15 / 0.951}   &  30.18 / 0.915  &  30.55 / 0.920\\
		MIMOUNetPlus+RIFE & 0.16 / 50.00 &  & 30.83 / 0.921 &  31.28 / 0.926 &    &  30.33 / 0.913
		& 30.69 / 0.918 \\
		
		UTI-VFI &  0.68 / 11.76  & ---  &  ---   & 34.40 / 0.965 & --- & ---   & 33.28 / 0.955 \\
		
		FLAVR &  0.50 / 16.00 & 35.78 / 0.968 & 34.20 / 0.959   &  34.39 / 0.960     &  33.72 / 0.953 & 33.36 / 0.952  & 33.41 / 0.952   \\
		
		DeMFI &  3.19 / 2.51 &36.69 / 0.974   & 34.77 / 0.965   & 35.01 / 0.967   & 35.01 / 0.964   &  34.43 / 0.962    & 34.50 / 0.962      \\
		\hline
		VIDUE (Ours) & 0.27 / 29.63 &\textbf{37.74 / 0.980} &  \textbf{36.12 / 0.973}   &  \textbf{36.32 / 0.974} &\textbf{36.12 / 0.973} &  \textbf{35.56 / 0.970}    & \textbf{35.63 / 0.970}       \\
		\hline
		
		\hline
	\end{tabular}
	\caption{{PSNR / SSIM comparison results on the GoPro dataset. ``Deblurring'' and ``Interpolation'' columns contain the reconstruction results of the input and interpolated frames, respectively, while the ``Avg'' column summarizes the average performance.}
	%Total\_
% 		* means the method is re-trained or finetuned by adopting the same training setting with our VIDUE.
	}
	\label{tabel:gopro8x}
\end{table*}
\begin{figure*}[!t] \footnotesize
	%	\hspace{-0.8cm}
	\setlength{\abovecaptionskip}{-2pt} 
	\setlength{\belowcaptionskip}{0pt}
	
	\flushbottom
	\begin{subfigure}[b]{0.15\linewidth} 
			\flushleft 
			\begin{tabular}{cccccc}
				\multirow{2}{*}{\includegraphics[width=0.85\linewidth]{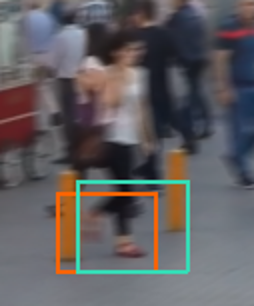}} \\
				\vspace{2.1cm}
				%				{Blurry Frame} \\  %\multirow{2}{*}
				%				{} \\
			\end{tabular}
			\centering
			\vspace{-4.7em}
			{\scriptsize Blurred Frame}\\
	\end{subfigure}
	%	\hspace{-1em}
		%\scriptsize
	\begin{subfigure}[b]{0.8\linewidth}
		%			\flushleft 			
		\begin{tabular}{ccccccccccc}
			%				\centering
			\hspace{-1.5em}
			\includegraphics[width=0.11\linewidth]{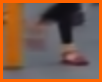} &
			\hspace{-1.2em}
			\includegraphics[width=0.11\linewidth]{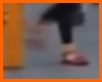} &
			\hspace{-1.2em}
			\includegraphics[width=0.11\linewidth]{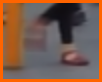} &
			\hspace{-1.2em}
			\includegraphics[width=0.11\linewidth]{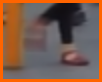} &
			\hspace{-1.2em}
			\includegraphics[width=0.11\linewidth]{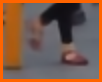} &
			\hspace{-1.2em}
			\includegraphics[width=0.11\linewidth]{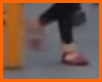} &
			\hspace{-1.2em}
			\includegraphics[width=0.11\linewidth]{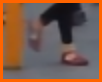} &
			\hspace{-1.2em}
			\includegraphics[width=0.11\linewidth]{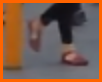} &
			\hspace{-1.2em}
			\includegraphics[width=0.11\linewidth]{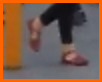}\\
			\hspace{-1.5em}
			\includegraphics[width=0.11\linewidth]{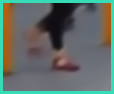} &
			\hspace{-1.2em}
			\includegraphics[width=0.11\linewidth]{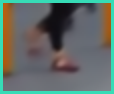} &
			\hspace{-1.2em}
			\includegraphics[width=0.11\linewidth]{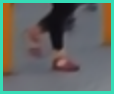} &
			\hspace{-1.2em}
			\includegraphics[width=0.11\linewidth]{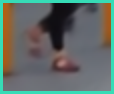} &
			\hspace{-1.2em}
			\includegraphics[width=0.11\linewidth]{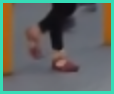} &
			\hspace{-1.2em}
			\includegraphics[width=0.11\linewidth]{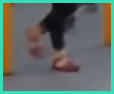} &
			\hspace{-1.2em}
			\includegraphics[width=0.11\linewidth]{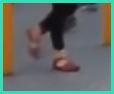} &
			\hspace{-1.2em}
			\includegraphics[width=0.11\linewidth]{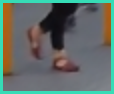} &
			\hspace{-1.2em}
			\includegraphics[width=0.11\linewidth]{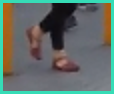}\\
			\hspace{-1.2em}
			{\scriptsize TSP + QVI} &
			\hspace{-1.2em}
			{\scriptsize TSP + RIFE} &
			\hspace{-1.2em}
			{\scriptsize MIMO + QVI} &
			\hspace{-1.2em}
			{\scriptsize MIMO + RIFE} &
			\hspace{-1.2em}
			{\scriptsize UTI-VFI} &
			\hspace{-1.2em}
			{\scriptsize FLAVR} &
			\hspace{-1.2em}
			{\scriptsize DeMFI} &
			\hspace{-1.2em}
			{\scriptsize VIDUE (Ours)} &
			\hspace{-1.2em}
			{\scriptsize GT} \\
		\end{tabular}
	\end{subfigure}
%	\vspace{-0.45em}
	\caption{Visual comparison on the GoPro dataset. 
		Orange and green boxes indicate cropped patches for zoomed-in comparison from deblurred and interpolated frames, respectively. TSP and MIMO are short for CDVD-TSP and MIMOUNetPlus, respectively. GT stands for the ground-truth patches cropped from the reference high framerate video.}
	\label{fig:gopro8x}
\end{figure*}
\noindent\textbf{Inter-Motion Analysis}. We adopt a second randomly initialized light-weight U-Net, taking the estimated intra-motion offsets $\bm{o}^{(s)}$ and $\bm{o}^{(e)}$ as inputs, and producing  inter-motion maps $\bm{n}\in\mathbb{R}^{(S\times T)\times 2\times H\times W}$. The adaptive gain tuning in Eq.~\eqref{eq:gaint} is also enabled in the expansive path of the U-Net.
The detailed network specifications of the motion analyzer can be found in the supplementary.

\subsection{Video Reconstruction Network}
It is then ready to describe our video reconstruction network $f: \mathbb{R}^{T\times 3\times H\times W}\mapsto\mathbb{R}^{(S\times T)\times 3\times H\times W}$, which is built upon the exposure-aware representation $\bm u$ and the motion-aware representations $\bm m$ and $\bm n$ by progressive exposure-adaptive convolution and motion refinement. The architecture is shown in Fig.~\ref{fig:framework}.

We first feed the input video sequence $\bm y\in\mathbb{R}^{(T\times3)\times H\times W}$ 
into an encoder, which consists of four stages of residual blocks, separated by the spatially downsampling layers. Each residual block contains two $3 \times 3$ convolution layers with leaky ReLU in between.
Similarly, the decoder is composed of three stages of one transposed convolution for upsampling, residual blocks, one exposure-adaptive convolution, and one motion refinement module, followed by a back-end residual block and a convolution layer for channel number adjustment. 

\noindent\textbf{Exposure-Adaptive Convolution}.
Motivated by~\cite{kang2017incorporating,karras2019style}, we propose to further convolve the intermediate representation $\bm z$ from the residual blocks using a filter bank, in which we adapt the filter weights to the exposure-aware representation $\bm u$. We generate the filter weights $\bm w$ by 
\begin{equation}\label{eq:eac}
	\begin{aligned}
		\bm{u}' &=\mathbf{W}_2^\intercal \mathrm{LReLU}\left(\mathbf{W}_1^\intercal \bm{u}\right),\\
		\bm{w}_{ijk} &= \frac{\bm{w}'_{ijk}\cdot \bm{u}'_i}{\sqrt{\sum_{i,k} (\bm{w}'_{ijk} \cdot \bm{u}'_i)^2 + \epsilon}},
	\end{aligned}
\end{equation}
where $\{\bm{w}',\mathbf{W}_1, \mathbf{W}_2\}$ are learnable, and $i$, $j$, and $k$ index the input feature channel, the output feature channel, and the spatial footprint of the convolution, respectively. $\epsilon$ is a small positive constant to avoid numerical instability issues. As in~\cite{karras2019style}, we first perform instance normalization of the feature representation $\bm z$ with the learnable scaling factor and the bias term, followed by the exposure-adaptive convolution to obtain $\bm g$ (refer to the notations in Fig.~\ref{fig:framework}).

\noindent\textbf{Motion Refinement Module}. At the $l$-th stage for $l\in\{1,2,3\}$, the motion refinement module accepts $\bm g^{(l)}$ from the exposure-adaptive convolution as input, whose channel dimension is adjusted by a $1\times 1$ convolution. The intra-motion maps $\bm{m}$ and the inter-motion maps $\bm{n}$ are accordingly resized to the same spatial size with $\bm{g}^{(l)}$, denoted by $\bm{m}^{(l)}$ and $\bm{n}^{(l)}$, respectively. We refine the intra-motion maps by 
\begin{align}
	\tilde{\bm{m}}^{(l)} = \sigma\left(\mathrm{Conv}\left(\mathrm{Concat}\left(\bm{m}^{(l)},\bm{g}^{(l)}\right)\right) + \bm{m}^{(l)}\right),
\end{align}
where $\mathrm{Conv}(\cdot)$ denotes a $3\times3$ convolution, $\mathrm{Concat}(\cdot)$ denotes the concatenation operation along the channel dimension, and $+$ indicates that we rely on a residual connection for implementation. We continue to refine  the inter-motion maps $\bm{n}^{(l)}$ based on $\tilde{\bm{m}}^{(l)}$:
\begin{equation}
	\tilde{\bm{n}}^{(l)}\!\!= \mathrm{Conv}(\mathrm{E\_Conv}(\bm{n}^{(l)}\!\!, \bm{u}) \odot \tilde{\bm{m}}^{(l)}\!\! + \!\bm{n}^{(l)} \odot (1\!-\!\tilde{\bm{m}}^{(l)})),
\end{equation}
where $\mathrm{E\_Conv(\cdot)}$ stands for the exposure-adaptive convolution and $\odot$ is an element-wise multiplication operation.

\noindent\textbf{Progressive Reconstruction}. Finally, we make use of the refined inter-motion maps $\tilde{\bm{n}}^{(l)}$ and the input features $\bm{g}^{(l)}$ for progressive reconstruction:
\begin{align}
	\hat{\bm{g}}^{(l)} = \mathrm{Refine}(\mathrm{Warp}(\bm{g}^{(l)}, \tilde{\bm{n}}^{(l)})) + \mathrm{Up}(\hat{\bm{g}}^{(l-1)}),
\end{align}
where $\mathrm{Warp}(\cdot)$ denotes the backward warping function~\cite{pan2020cascaded}, and $\mathrm{Up}(\cdot)$ denotes the bilinear upsampling operation. $\mathrm{Refine}(\cdot)$ is implemented by a front-end convolution layer, two residual blocks, and a back-end convolution layer. To initialize progressive reconstruction, we set $\hat{\bm{g}}^{(0)} = \bm{0}$ as a tensor with all zeros, and summarize one stage of processing in Fig.~\ref{fig:g_a}. At last, we add $\bm{g}^{(4)}$ (\ie, the output of the back-end residual block and convolution layer in the decoder) to  $\hat{\bm{g}}^{(3)}$ to estimate the high framerate sharp video sequence $\hat{\bm{x}}\in\mathbb{R}^{(S\times T)\times 3\times H\times W}$.

During training, we optimize all modules in the proposed VIDUE method (except for the exposure-aware feature extractor) using a variant of stochastic gradient descent by minimizing the Mean Absolute Error (MAE) between the ground-truth high framerate sharp video sequences and their predictions by VIDUE.
%
%Then we transfer the video restoration under arbitrary exposure time into a video restoration conditioned on the exposure time setting.

%
% As shown in Eq. \eqref{eq:loss}, our VIDUE consists of two parts: exposure setting predictor $\mathcal{T}$ and video restoration network conditioned on temporal condition $\bm{v}$.
%

%

%$\bm{v}$ is an implicit representation of exposure time setting of the input sequence, and exposure time setting has a positive correlation on the blur degree of the sequence, which can be regarded as temporal style of the sequence.
% Exposure representation $\bm v$ can be embedded into conditional reconstruction network $\mathcal F$ by simply applying Channel Attention (CA)~\cite{hu2018squeeze}. Yet, 

\begin{table*}[t]\footnotesize  %[!htb]\small  %[t]  %
	\centering
	\setlength{\abovecaptionskip}{0pt} 
	\setlength{\belowcaptionskip}{0pt}
	\begin{tabular}{l|c|c|c|c|c|c}
		\hline
		
		\hline	
		\multirow{2}{*}{Method}  &\multicolumn{3}{c|}{Adobe-5:8}  &  \multicolumn{3}{c}{Adobe-7:8} \\
		\cline{2-7}
		& Deblurring  & Interpolation & Avg & Deblurring  & Interpolation & Avg\\
		\hline
		CDVD-TSP+QVI & \multirow{2}{*}{29.34 / 0.896} & 23.34 / 0.739   & 24.10 / 0.759    &   \multirow{2}{*}{27.57 / 0.862}   & 23.42 / 0.741  &  23.94 / 0.756  \\
		CDVD-TSP+RIFE &  & 23.29 / 0.729   &  24.06 / 0.750   &     &  23.31 / 0.730 &  23.85 / 0.747  \\
% 		\hline
		MIMOUNetPlus+QVI  &\multirow{2}{*}{29.71 / 0.894} &  23.48 / 0.741
		& 24.27 / 0.761  &  \multirow{2}{*}{27.69 / 0.854}   &  23.52 / 0.740  &  24.05 / 0.754 \\
		MIMOUNetPlus+RIFE &  & 23.85 / 0.748 & 24.59 / 0.767  &  & 23.76 / 0.744	& 24.26 / 0.758 \\
% 		\hline
		UTI-VFI & ---  & ---  & 26.69 / 0.842  & ---  & ---  & 25.84 / 0.815 \\
% 		\hline
		FLAVR & 28.52 / 0.869  & 27.05 / 0.841  &  27.23 / 0.845  & 26.98 / 0.834  & 26.94 / 0.837  & 26.94 / 0.837   \\
% 		\hline
		DeMFI &  27.24 / 0.848  & 25.49 / 0.814   &  25.71 / 0.818  &  25.81 / 0.813  &  25.64 / 0.814    & 25.66 / 0.814      \\
		\hline
		VIDUE (Ours) & \textbf{30.44 / 0.905} &  \textbf{28.50 / 0.876}   &  \textbf{28.74 / 0.880} &\textbf{27.85 / 0.864} &  \textbf{27.92 / 0.867}    & \textbf{27.91 / 0.866}       \\
		\hline
		
		\hline
	\end{tabular}
	\caption{PSNR / SSIM comparison results on the Adobe dataset. }
	\label{tabel:adobe8x}
\end{table*}
\begin{figure*}[t] \footnotesize
	%	\hspace{-0.8cm}
	\setlength{\abovecaptionskip}{-2pt} 
	\setlength{\belowcaptionskip}{0pt}
	\hspace{-0.2cm}
	\flushbottom
		\begin{subfigure}[b]{0.17\linewidth} 
			%			\flushleft 
			\hspace{-2em}
			\begin{tabular}{cccccc}
				\multirow{2}{*}{\includegraphics[width=0.95\linewidth]{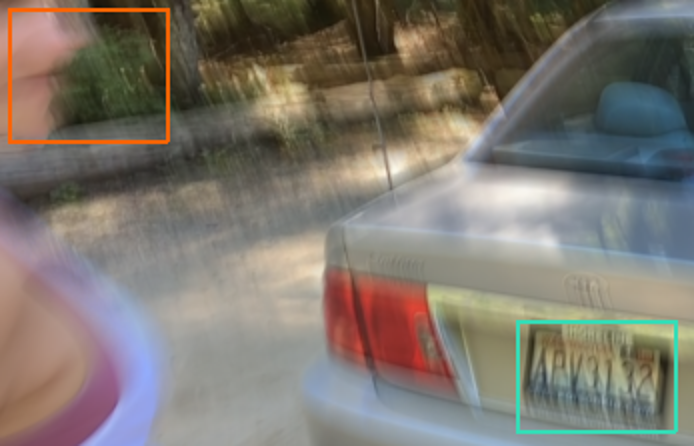}} \\
				\vspace{1.4cm}
				%				{Blurry Frame} \\  %\multirow{2}{*}
				%				{} \\
			\end{tabular}
			\centering
			{\scriptsize Blurred Frame}
			\vspace{-4em}
	\end{subfigure}
%	\hspace{-0.5em}
		%\scriptsize
		\begin{subfigure}[b]{0.8\linewidth}
			%			\flushleft 			
			\begin{tabular}{ccccccccccc}
				%				\centering
				\hspace{-2em}
				\includegraphics[width=0.11\linewidth]{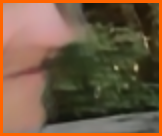} &
				\hspace{-1.2em}
				\includegraphics[width=0.11\linewidth]{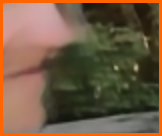} &
				\hspace{-1.2em}
				\includegraphics[width=0.11\linewidth]{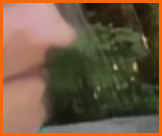} &
				\hspace{-1.2em}
				\includegraphics[width=0.11\linewidth]{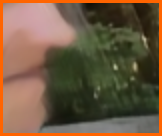} &
				\hspace{-1.2em}
				\includegraphics[width=0.11\linewidth]{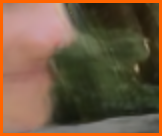} &
				\hspace{-1.2em}
				\includegraphics[width=0.11\linewidth]{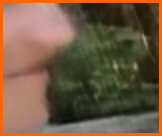} &
				\hspace{-1.2em}
				\includegraphics[width=0.11\linewidth]{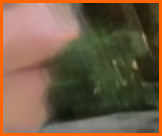} &
				\hspace{-1.2em}
				\includegraphics[width=0.11\linewidth]{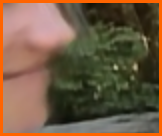} &
				\hspace{-1.2em}
				\includegraphics[width=0.11\linewidth]{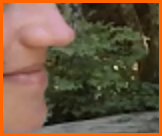}\\
				\hspace{-2em}
				\includegraphics[width=0.11\linewidth]{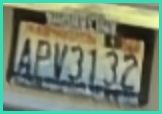} &
				\hspace{-1.2em}
				\includegraphics[width=0.11\linewidth]{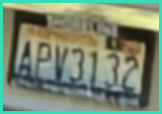} &
				\hspace{-1.2em}
				\includegraphics[width=0.11\linewidth]{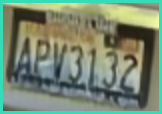} &
				\hspace{-1.2em}
				\includegraphics[width=0.11\linewidth]{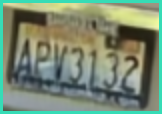} &
				\hspace{-1.2em}
				\includegraphics[width=0.11\linewidth]{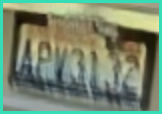} &
				\hspace{-1.2em}
				\includegraphics[width=0.11\linewidth]{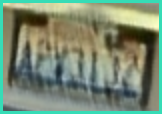} &
				\hspace{-1.2em}
				\includegraphics[width=0.11\linewidth]{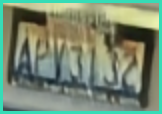} &
				\hspace{-1.2em}
				\includegraphics[width=0.11\linewidth]{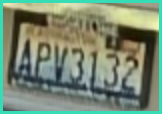} &
				\hspace{-1.2em}
				\includegraphics[width=0.11\linewidth]{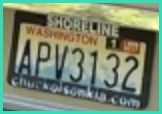}\\
				\hspace{-1.5em}
				{\scriptsize TSP + QVI} &
				\hspace{-1.2em}
				{\scriptsize TSP + RIFE} &
				\hspace{-1.2em}
				{\scriptsize MIMO + QVI} &
				\hspace{-1.2em}
				{\scriptsize MIMO + RIFE} &
				\hspace{-1.2em}
				{\scriptsize UTI-VFI} &
				\hspace{-1.2em}
				{\scriptsize FLAVR} &
				\hspace{-1.2em}
				{\scriptsize DeMFI} &
				\hspace{-1.2em}
				{\scriptsize VIDUE (Ours)} &
				\hspace{-1.2em}
				{\scriptsize GT} \\
			\end{tabular}
	\end{subfigure}
%	\vspace{-0.45em}
	\caption{Visual comparison on the Adobe dataset. 
		Orange and green patches are from deblurred and interpolated frames, respectively. }
	\label{fig:adobe8x}
\end{figure*}

\section{Experiments}
%In this section, we first introduce datasets and implementation details. Then we conduct ablation studies to analyze the contribution of each proposed component. Finally, we compare the proposed model with state-of-the-art algorithms.

In this section, we evaluate VIDUE on both synthetic and real-world datasets. More results including reconstructed videos can be found in the supplementary. 
The source code is implemented in Pytorch, and is made publicly available at {\url{https://github.com/shangwei5/VIDUE}}. We also provide an implementation in HUAWEI Mindspore at \url{https://github.com/Hunter-Will/VIDUE-mindspore}.

\vspace{-0.2em}
\subsection{Datasets}
To establish simulated datasets for quantitative performance comparison of video interpolation and deblurring, we synthesize low framerate videos according to  Eq.~\eqref{eq:gen} by downsampling high framerate videos~\cite{Oh2022DeMFI,zhang2020video}. 
The experiments are conducted on both the GoPro dataset~\cite{Nah_2017_CVPR} and the Adobe dataset~\cite{su2017deep}. 
% ground-truth frames as done in . 
%
%The framerate of the ground-truth sequence is $\Delta t$ times that of the blurry sequence. We discard $\Delta t - t_e = t_r$ video frames to simulate the readout time. 
%
On the GoPro dataset with $\times 8$ interpolation (and deblurring) task, we set $S = 8$ by which an original 240 FPS video is degraded to 30 FPS. 
% which is a common FPS setting in our daily life.
%
For a fair comparison, we set exposure frames to two odd numbers, \ie, $\Delta t_e : \Delta t\in \{5:8, 7:8\}$, since existing methods require the middle frame as reference. 
%$\Delta t_e : \Delta t = S:S$
 As for training, we sample $\Delta t_e : \Delta t \in\{ 1:8, 2:8, \ldots, 8:8\}$ to generate blurred frames as a form of data augmentation to  train the exposure-aware feature extractor $g_e$.
%
%With more kinds of exposure time data can help our predictor $\mathcal{T}$ learning representation vector better.
%
%  and name these synthetic datasets as "dataset-$t_e$-$t_r$". 
On the Adobe dataset with $\times 8$ interpolation (and deblurring) task, the synthetic setting is identical, but the blurring artifacts are more severe by setting a longer shutter period.
% we magnify $t_e$ and $\Delta t$ by three times to generate video data with a high degree of blur. We use these datasets under different exposure times and different blur degree to verify the effectiveness of our method. 
%
%Finally we get "GoPro-5-3", "GoPro-7-1" with slight blur and "Adobe-5-3", "Adobe-7-1" with heavy blur respectively.
%%
%In addition, we also provide datasets "GoPro-9-7", "GoPro-11-5", "GoPro-13-3" and "GoPro-15-1" to evaluate our method on MFI(x16).
%
Furthermore, we evaluate the generalizability of VIDUE against competing methods on real-world data from the RealBlur dataset~\cite{su2017deep}.

\begin{table*}[t] \footnotesize %[!htb]\small
	\setlength{\tabcolsep}{12pt}
	\setlength{\abovecaptionskip}{0pt} 
	\setlength{\belowcaptionskip}{0pt}
	\centering
	\begin{tabular}{l|c|c|c|c|c}
		\hline
		
		\hline	
		{Method}  &{GoPro-9:16}  & {GoPro-11:16} & {GoPro-13:16} & {GoPro-15:16} & {Avg} \\  %\midrule[1pt]
		\hline
		CDVD-TSP+QVI & 25.70 / 0.786 & 25.45 / 0.780 &  25.20 / 0.774  & 24.97 / 0.767 & 25.33 / 0.777 \\
		CDVD-TSP+RIFE & 26.08 / 0.797 & 25.74 / 0.788   &  25.41 / 0.780   &  25.12 / 0.772 & 25.59 / 0.784 \\
% 		\hline
		MIMOUNetPlus+QVI  & 26.12 / 0.794  & 25.97 / 0.791 & 25.78 / 0.787  & 25.49 / 0.780 &25.84 / 0.788 \\
		MIMOUNetPlus+RIFE & 26.95 / 0.821 & 26.71 / 0.815 & 26.41 / 0.808  &  25.98 / 0.797 & 26.51 / 0.810\\
% 		\hline
		UTI-VFI & 28.59 / 0.876  & 28.17 / 0.866  & 27.40 / 0.845 & 26.37 / 0.814  & 27.63 / 0.851 \\
% 		\hline
		FLAVR & 28.77 / 0.875  & 28.71 / 0.875   &  28.43 / 0.870   &28.04 / 0.860    & 28.49 / 0.870  \\
% 		\hline
		DeMFI & 28.64 / 0.876 & 28.67 / 0.878 & 28.49 / 0.873 & 28.11 / 0.863   & 28.48 / 0.873 \\
		\hline	
		%		Ours & \textbf{29.172/0.8942} &  \textbf{29.688/0.9045}   &  28.326/\textbf{0.8818} &27.963/\textbf{0.8702}  & \textbf{28.7873/0.8877}  \\
		VIDUE (Ours) & \textbf{31.10 / 0.922} &  \textbf{29.91 / 0.906}   &  \textbf{29.90 / 0.905} &\textbf{29.50 / 0.898}  & \textbf{30.10 / 0.908}  \\
		\hline
		
		\hline
	\end{tabular}
	\caption{PSNR / SSIM comparison results on the GoPro dataset for joint $\times 16$ interpolation and deblurring. }
	\label{tabel:GoPro16x}
\end{table*}
\begin{table}[!ht]\footnotesize  %[!htb]\small
	\setlength{\tabcolsep}{8pt}
	\centering
	\setlength{\abovecaptionskip}{0pt} 
	\setlength{\belowcaptionskip}{-5pt}
	\begin{tabular}{l|c|c}
		\hline
		
		\hline	
		Setting & GoPro-5:8  & GoPro-7:8 \\
		\hline
		Exposure-Agnostic & 34.79 / 0.965   &  34.13 / 0.960     \\
		Ordinal Regression & 35.44 / 0.969    & 34.97 / 0.966       \\
		\textbf{Contrastive Learning} & 36.32 / 0.974    & 35.63 / 0.970       \\
		\hline
		Known Exposure Time &36.41 / 0.974    & 35.87 / 0.971  \\
		\hline
		
		\hline
	\end{tabular}
	\caption{Role of the exposure-aware feature representation $\bm u$ evaluated by PSNR / SSIM. The default setting is highlighted in bold.}
	\label{tabel:tp}
\end{table}

\begin{table*}[!ht]\footnotesize  %[!htb]\small
	\setlength{\tabcolsep}{15pt}
	\centering
	\setlength{\abovecaptionskip}{0pt} 
	\setlength{\belowcaptionskip}{0pt}
	\begin{tabular}{c|c|c|c|c|cc}
		\hline
		
		\hline	
		\multicolumn{2}{c|}{Exposure Representation}  &  \multirow{1}{*}{Intra-}  &  Inter- &  Motion  & \multirow{2}{*}{GoPro-5:8}  & \multirow{2}{*}{GoPro-7:8} \\
		\cline{0-1}
		Squeeze-and-Excitation  & $\mathrm{E\_Conv}$ & Motion &Motion & Refinement & & \\
		\hline
		\XSolidBrush &\XSolidBrush &  \XSolidBrush  &  \XSolidBrush   &    \XSolidBrush   &  34.79 / 0.965     &   34.13 / 0.960   \\
		\XSolidBrush &\Checkmark &  \XSolidBrush  &  \XSolidBrush   &    \XSolidBrush   &  36.05 / 0.972     &   34.87 / 0.963   \\
		\XSolidBrush &\Checkmark &  \XSolidBrush  &  \Checkmark   &    \XSolidBrush   &  36.18 / 0.973   &  35.27 / 0.968   \\
		\XSolidBrush &\Checkmark &  \Checkmark &    \XSolidBrush         &   \Checkmark      &  36.20 / 0.973  &  35.24 / 0.968      \\
		\XSolidBrush &\Checkmark &  \Checkmark &    \Checkmark         &   \XSolidBrush       &   36.23 / \textbf{0.974}  &  35.28 / 0.968      \\
		\Checkmark &\XSolidBrush &  \Checkmark   &   \Checkmark     &   \Checkmark  &  36.12 / 0.973  & 35.39 / 0.969     \\
		\XSolidBrush & \Checkmark &  \Checkmark    &  \Checkmark  &\Checkmark   &  \textbf{36.32 / 0.974}    & \textbf{35.63 / 0.970}       \\
		
		\hline
		
		\hline
	\end{tabular}
	\caption{ Module analysis of VIDUE on the GoPro dataset. }
	\label{tabel:ablation}
\end{table*}

\begin{figure*}[!t] \footnotesize
	\setlength{\abovecaptionskip}{-2pt} 
	\setlength{\belowcaptionskip}{0pt}
	\hspace{-0.2cm}
	\flushbottom
		\footnotesize
		\begin{subfigure}[b]{0.15\linewidth} 
			\flushleft 
			\begin{tabular}{cccccc}
				\multirow{2}{*}{\includegraphics[width=0.95\linewidth]{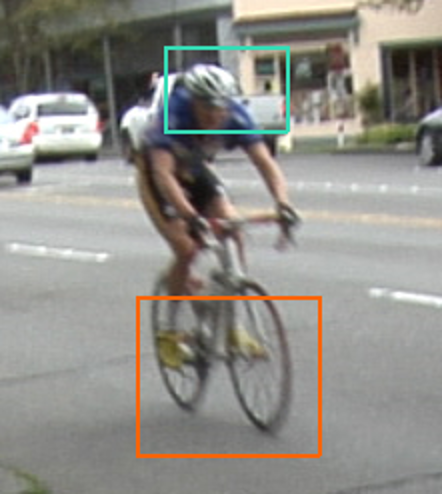}} \\
				\vspace{2.2cm}
				%				{Blurry Frame} \\  %\multirow{2}{*}
				%				{} \\
			\end{tabular}
			\centering
% 			\vspace{1em}
			{\scriptsize Blurred Frame}
			 			\vspace{-5em}
	\end{subfigure}
	\hspace{-0.5em}
		%\scriptsize
		\footnotesize
		\begin{subfigure}[b]{0.8\linewidth}
			%			\flushleft 			
			\begin{tabular}{ccccccccccc}
				%				\centering
				\hspace{-0.5em}
				\includegraphics[width=0.125\linewidth]{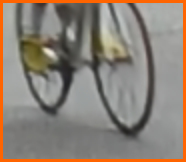} &
				\hspace{-1.2em}
				\includegraphics[width=0.125\linewidth]{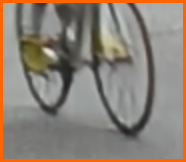} &
				\hspace{-1.2em}
				\includegraphics[width=0.125\linewidth]{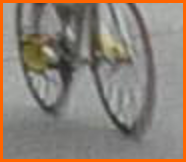} &
				\hspace{-1.2em}
				\includegraphics[width=0.125\linewidth]{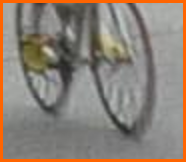} &
				\hspace{-1.2em}
				\includegraphics[width=0.125\linewidth]{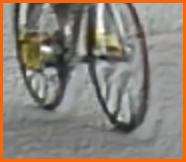} &
				\hspace{-1.2em}
				\includegraphics[width=0.125\linewidth]{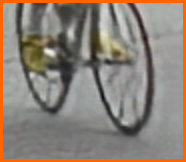} &
				\hspace{-1.2em}
				\includegraphics[width=0.125\linewidth]{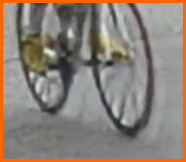} &
				\hspace{-1.2em}
				\includegraphics[width=0.125\linewidth]{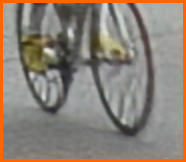} \\
				\hspace{-0.5em}
				\includegraphics[width=0.125\linewidth]{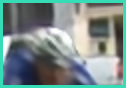} &
				\hspace{-1.2em}
				\includegraphics[width=0.125\linewidth]{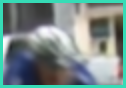} &
				\hspace{-1.2em}
				\includegraphics[width=0.125\linewidth]{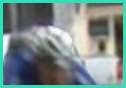} &
				\hspace{-1.2em}
				\includegraphics[width=0.125\linewidth]{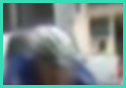} &
				\hspace{-1.2em}
				\includegraphics[width=0.125\linewidth]{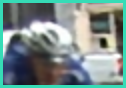} &
				\hspace{-1.2em}
				\includegraphics[width=0.125\linewidth]{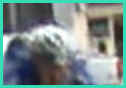} &
				\hspace{-1.2em}
				\includegraphics[width=0.125\linewidth]{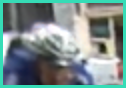} &
				\hspace{-1.2em}
				\includegraphics[width=0.125\linewidth]{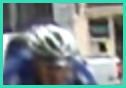} \\
				\hspace{-0.5em}
				{\scriptsize TSP + QVI} &
				\hspace{-1.2em}
				{\scriptsize TSP + RIFE} &
				\hspace{-1.2em}
				{\scriptsize MIMO + QVI} &
				\hspace{-1.2em}
				{\scriptsize MIMO + RIFE} &
				\hspace{-1.2em}
				{\scriptsize UTI-VFI} &
				\hspace{-1.2em}
				{\scriptsize FLAVR} &
				\hspace{-1.2em}
				{\scriptsize DeMFI} &
				\hspace{-1.2em}
				{\scriptsize VIDUE (Ours)}\\
			\end{tabular}
	\end{subfigure}
%	\vspace{-0.4em}
	\caption{Visual comparison on the RealBlur dataset. Orange and green patches are from
		deblurred and interpolated frames, respectively.}
	\label{fig:real}
\end{figure*}

\vspace{-0.2em}
\subsection{Implementation Details}
We set the input frame number $T = 4$, and the the exposure-aware feature dimension $C = 256$. The temperature parameter in Eq.~\eqref{eq:contr} and the normalizing constant in Eq.~\eqref{eq:eac} is set to $\alpha = 0.5$ and $\epsilon = 1 \times 10^{-5}$, respectively. We adopt the Adam~\cite{kingma2014adam} optimizer with the default setting for training $g_e$ with an initial learning rate of $0.1$ and a mini-batch size of $40$. Similarly, we adopt the AdaMax~\cite{kingma2014adam} optimizer with parameters $\beta_1 = 0.9$ and $\beta_2 = 0.999$ for training $g_a$ and $f$, and set the initial learning rate and the mini-batch size to $2 \times 10^{-4}$ and $12$, respectively.
We train the models for $200$ epochs, and halve the learning rate whenever the training plateaus, which is cross-validated as done in~\cite{kalluri2023flavr}. VIDUE is trained on $4$ Tesla V100 GPUs, and can make inference on a single GTX 2080 Ti GPU.
 Throughout the paper, we use the Peak Signal-to-Noise Ratio (PSNR) and the Structural SIMilarity (SSIM) index~\cite{wang2004image} as the evaluation metrics.

\vspace{-0.2em}
\subsection{Evaluation on $\times 8$ Interpolation Task}
We compare VIDUE with both cascade and joint methods. 
% We mainly compare our VIDUE with both cascade scheme methods and joint scheme methods. 
For cascade methods, 
% we connect deblurring and interpolation models, including
deblurring methods CDVD-TSP~\cite{pan2020cascaded}, MIMOUNetPlus~\cite{cho2021rethinking} and interpolation methods QVI~\cite{xu2019quadratic}, RIFE~\cite{huang2022rife} are cascaded for video deblurring and interpolation. We also take UTI-VFI~\cite{zhang2020video} into comparison, which handles unknown exposure time.
For joint methods, we include FLAVR~\cite{kalluri2023flavr} and DeMFI~\cite{Oh2022DeMFI}. 
It is worth noting that the original FLAVR is an interpolation method with sharp inputs, and we retrain it on the same training sets to tackle joint video interpolation and deblurring. 
% , dubbed FLAVR*
% We also retrain or finetune the other competing methods on datasets for fair comparison. 
%
For MIMOUNetPlus and RIFE, we find that models provided by the respective authors perform better than our retrained counterparts, and thus we stick to the official implementations for evaluation. 
%
% We note that ALANET~\cite{gupta2020alanet} and BIN~\cite{shen2020blurry} are not taken into comparison, because they can only interpolate the center frame between blurred input sequence, whose quantitative metrics cannot be consistent with the competing methods. 
%
%\subsubsection{GoPro}
%The GoPro high speed video dataset~\cite{nah2017deep}, a benchmark for dynamic scene deblurring, provides 33 720P videos taken at 240fps. We used 22 videos for training and 11 for testing. We apply the synthetic rule on GoPro dataset~\cite{nah2017deep},
%
%%
%\subsubsection{Adobe}
%We also use the Adobe240 dataset~\cite{su2017deep} for training and testing. It consists of 120 videos at 240 fps with the resolution of 1280 $\times$ 720. We use 112 of the videos to construct the training set.
%
%%
%\subsubsection{RealBlur}
%To evaluate the generalization of our method on real blurry videos, we use the real video deblurring dataset(RealBlur)\cite{su2017deep} for generalization experiments.

\subsubsection{Comparison on the GoPro Dataset}
% On GoPro dataset, we evaluate the deblurring performance and multi-frame interpolation (x8) performance in Table \ref{tabel:gopro8x}. 
%
As mentioned previously, we create two versions of the GoPro dataset, ``GoPro-5:8" and ``GoPro-7:8", with different exposure time, and list the comparison results in Table~\ref{tabel:gopro8x}. We find that cascade methods are significantly worse than joint methods, with a PSNR gap of almost $3$ to $6$ dB. 
%
%That is because the interpolation performance is highly dependent on the deblurred frames.
%
In comparison with joint methods, VIDUE achieves about $1.0$ to $1.1$ dB PSNR gains for deblurring, and about $1.1$ to $1.4$ dB PSNR gains for interpolation, respectively.
%
%This is because VIDUE can benefit from learned temporal condition, which can condition the video restoration according to exposure time. 
%
%Moreover, our motion analysis module can compensate the motion shift in spatial dimension according to calculated motion map.
%
In terms of visual comparison in Fig.~\ref{fig:gopro8x}, cascade methods fail in interpolating sharp frames. 
 The deblurring results of the joint methods FLAVR and DeMFI are satisfactory, but they are still limited in interpolation. 
This is because the unknown exposure time setting is not carefully modeled, and latent sharp frames during the readout phase cannot be properly reconstructed. 
% times  in training data make influence on them, and they can only get average results instead conditioning restoration according to different exposure time.
%
In stark contrast, VIDUE is able to adapt to different exposure time based on the learned exposure-aware feature representation, and achieves noticeable performance gains on this challenging task of joint video multi-frame interpolation and deblurring. 
In addition, we test the inference time (and framerate) of all methods on the GoPro dataset.
% It is worth noting that some interpolation methods can only generate one frame at one time, while others can generate multiple frames at one time. For fair comparison, 
We calculate the average running time of reconstructing $8$ frames (\ie, using $\times 8$ interpolation as the example) on the Tesla V100 GPU.
We find from Table~\ref{tabel:gopro8x} that VIDUE shows clear advantages over all competing methods except for the light-weight cascade method (\ie, MIMOUNetPlus+RIFE, which does not deliver convincing reconstruction performance). 
DeMFI involves extensive recurrent computation, leading to the longest inference time among all methods. 
In summary, the proposed VIDUE enjoys faster inference time, and achieves the best reconstruction performance, which justifies our key design philosophy of adaptive computation to deblurring and interpolation relevant features.
% style, which leads to a optimal result of different settings.
%
% Above all, our VIDUE can outperform both joint scheme methods and cascade scheme methods, and achieve pleasure resluts in Fig. \ref{fig:gopro8x}.

\subsubsection{Comparison on the Adobe Dataset}
%In order to prove that our method also has advantages in the case of heavy blur, we create heavy blur dataset with different exposure time, "Adobe-5-3" and "Adobe-7-1".
%
We evaluate the joint $\times 8$ interpolation and deblurring performance of VIDUE against the competing methods in Table~\ref{tabel:adobe8x} and Fig.~\ref{fig:adobe8x}.
Despite more severe blurring artifacts than those in the GoPro dataset, we come to similar conclusions that joint methods are superior over cascade methods. 
The most competitive method - DeMFI - experiences a significant performance drop due to the presence of the heavy blur.
VIDUE still outperforms FLAVR about $0.9$ to $1.9$ dB for deblurring, and about $1.0$ to $1.4$ dB for interpolation, respectively.
From Fig.~\ref{fig:adobe8x}, one can see that the competing methods fail to restore sharp frames, while VIDUE still obtains visually favorable results even in the presence of the strong blur. 

\subsubsection{Generalization to Real-World Videos}
Finally, we evaluate the generalizability of VIDUE on real-world blurred frames by testing the model trained on the GoPro dataset to interpolate and deblur video data from the RealBlur dataset. 
% for a fair comparision.
%  
As shown in Fig.~\ref{fig:real}, VIDUE achieves the most visually plausible interpolation and deblurring results with sharper structures and textures, while others suffer from visually annoying artifacts.
%
% Our method can generate least artifacts and sharper textures comparing with other methods. 
%
\subsection{Evaluation on $\times16$ Interpolation Task}
Without auxiliary information (\eg, event signals~\cite{kim2021event}), joint $\times 16$ interpolation and deblurring is extremely difficult, and is rarely evaluated in literature. 
We conduct this experiment on the GoPro dataset, and list the results in Table~\ref{tabel:GoPro16x}.  VIDUE achieves more than $1.5$ dB gains than existing methods, and performs consistently better under all exposure time settings, which are unknown. 
% cascade methods obtain inferior results due to the high difficulty of $\times 16$ interpolation. 
% Our VIDUE One can see our method can achieve higher PSNR/SSIM than both joint scheme methods and cascade scheme methods.
%
More visual comparisons are provided in the supplementary. 

% \begin{figure}[!t]\footnotesize
% 	\centering
% 	\setlength{\tabcolsep}{0pt}
% 	\setlength{\abovecaptionskip}{0pt} 
% 	\setlength{\belowcaptionskip}{0pt}
% 	\begin{tabular}{ccc}
% 		\includegraphics[width=0.33\linewidth]{ablation/tp/od}&
% 		\includegraphics[width=0.33\linewidth]{ablation/tp/ws}&
% 		\includegraphics[width=0.33\linewidth]{ablation/tp/odws}
% 		\vspace{-0.1in}
% 		\\
% 		{$\mathcal{L}_\text{od}$} &
% 		{$\mathcal{L}_\text{ws}$} &
% 		{$\mathcal{L}_\text{od} + \mathcal{L}_\text{ws}$}  \\
% 	\end{tabular}
% 	\caption{t-SNE visualization of learned temporal conditions $\bm v$ for $\times 8$ interpolation under different exposure time. Different colors correspond to exposure time $v \propto$ 1:8, 3:8, 5:8, 7:8.}
% 	\label{fig:tsne}
% \end{figure}  %Supplementary

\subsection{Ablation Studies}
\subsubsection{Effectiveness of Exposure-Aware Representation}
%We develop a predictor $\mathcal{T}$ to learn exposure time setting implicitly so that it is possible to restore video under arbitrary exposure time. 
%
As shown in Table~\ref{tabel:tp}, four VIDUE variants are trained on the GoPro dataset to verify the effectiveness of the exposure-aware feature representation $\bm u$: 1) one that is exposure-agnostic (\ie, without learning and adapting to $\bm u$), 2) one that learns $\bm u$ using ordinal regression, 3) one that learns $\bm u$ using supervised contrastive learning combined with relative weighting (as the default setting), 4) one with the known exposure time (as a form of upper bound). 
%
% To demonstrate the effectiveness of exposure-aware feature representation $\bm{u}$, we discard the exposure-aware feature representation $\bm{u}$, and retrain the method without using any exposure time representation. 
To leverage the known exposure time, we first represent it as an $S$-dimensional vector with
the first $\Delta t_e$ entries being one and the remaining entries being zero.  We then use two FC layers with leaky ReLU in between to map it into a $C$-dimensional feature representation, which can be readily adapted in VIDUE.
%
% \red{Moreover, we train reconsturction network with exposure-aware representation learned by $\ell_{od}$. }
%
% And we also analyzed the influence of different losses on the capability of predictor $\mathcal{T}$. 
% We train all these variants with the same setting for fair comparison on GoPro dataset.
%
% Specifically, we used GoPro to generate low framerate blurry frames with different exposure time settings and feed them to the predictor $\mathcal{T}$ training with only $\mathcal{L}_\text{od}$ and $\mathcal{L}_\text{od} + \mathcal{L}_\text{ws}$ to produce temporal conditions. 
%
% The learned temporal conditions are visualized using t-SNE~\cite{van2008visualizing} in Fig. \ref{fig:tsne}.
%
% One can see that training with only $\mathcal{L}_\text{od}$ cannot distinguish sequence with various exposure time, and training with $\mathcal{L}_\text{ws}$ can obtain discriminative representations $\bm u$. Finally, combining $\mathcal{L}_\text{od}$ with $\mathcal{L}_\text{ws}$ can achieve best results. 
% Combining with 2$ \sim $3 rows in Table \ref{tabel:tp}, we note that the discrimination of temporal condition is positively correlated with the quality of video restoration.
%
%Table \ref{tabel:tp} and Fig. \ref{fig:tp} show both quantitative and qualitative evaluations for using different temporal conditions. 
%
% For the task under unknown exposure time, it is most crucial to distinguish different exposure time. 
As reported in Table~\ref{tabel:tp}, both exposure-aware feature representations learned by ordinal regression and supervised contrastive learning bring significant improvements than the exposure-agnostic variant. Compared to ordinal regression, our default choice of supervised contrastive learning is able to approach the ``upper bound'' with the known exposure time. We believe this arises because contrastive learning provides a more direct way of encouraging discriminative feature learning than ordinal regression learns to rank different exposure time.

% \red{Comparing with the ``w/o Exposure Time" variant, exposure-aware feature representations learned by $\ell_{od}$ and $\ell_{ws}$ bring significant improvement in Table ~\ref{tabel:tp}. 
% %
% For this task, it is most crucial to distinguish different exposure time, and $\ell_{od}$ brings extra training difficulty to exactly predict exposure time. Therefore, we finally choose $\ell_{\mathrm{ws}}$, which can approach the upper bound by providing the accurate exposure time.}
%
% One can see that our method with $\ell_{\mathrm{ws}}$ is approaching the upper bound by providing the accurate exposure time . 
% Comparing with the ``w/o Exposure Time" variant, our method can achieve about \emph{+1.5 dB} PSNR gain under different exposure time.
% We note that our method can reach the upper bound with known exposure time settings, and our method with $\mathcal{L}_\text{od} + \mathcal{L}_\text{ws}$ is also close to the upper bound.
% Qualitative evaluations can be found in supplementary. 
%
%Fig. \ref{fig:tp} further shows that using more discriminative representations is able to generate the frames with clearer structures. %Supplementary
%
% Moreover, facing videos with unknown exposure time settings in reality, our method can also implicitly learn the exposure time settings to condition the video restoration. 
\subsubsection{Module Analysis}
We single out the contribution of each component of VIDUE using the GoPro dataset. 
The first row in Table~\ref{tabel:ablation} shows the results of a plain U-Net, while the second row is obtained by a simplified VIDUE without motion analysis and adaptation in the decoder. Surprisingly, the simplified VIDUE achieves better results than existing methods, (see Tables~\ref{tabel:gopro8x} and~\ref{tabel:ablation}).
 The performance gains by VIDUE is mostly attributed to the use of the exposure-aware representation by contrasting it to the plain U-Net.
We next remove each component of VIDUE to verify its necessity. 
Most importantly, removing motion refinement leads to performance drops under different exposure time, especially when the exposure time is large (\ie, when the motion is strong).
We also replace the exposure-adaptive convolution (\ie, $\mathrm{E\_Conv}$ defined in Eq.~\eqref{eq:eac}) with the sequence-and-excitation operation~\cite{hu2018squeeze}, and find that 
$\mathrm{E\_Conv}$ is more effective in leveraging the exposure-aware representation.
As expected, the full VIDUE achieves the best interpolation and deblurring performance.
\section{Conclusion}
We have described a computational method - VIDUE - for joint video multi-frame interpolation and deblurring under unknown exposure time. 
We trained contrastively to extract exposure-aware feature representation, which can then be embedded into intra- and inter-motion analyzer and the video reconstruction network via gain tuning and exposure-adaptive convolution, respectively. We refined the estimated motion representations for better progressive video reconstruction.
% In this way, our method can well handle low framerate blurred videos with different exposure time to reconstruct high framerate sharp videos.
We demonstrated the superiority of VIDUE on both synthetic and real-world datasets to perform $\times 8$ and $\times 16$ interpolation and deblurring tasks. 
% Especially on more difficult $\times 16$ interpolation task, our method can still achieve more than \emph{1.5 dB} PSNR gain for joint video deblurring and interpolation.  
%
Future work can be planed to 1) further reduce the computational complexity of VIDUE while retaining (or improving) the performance and 2) to optimize VIDUE by perceptual quality metrics with emphasis on temporal coherence. 

\section*{Acknowledgements}
This work was supported in part by the National Key Research and Development Project (2022YFA1004100), the National Natural Science Foundation of China (62172127, 62071407, and U22B2035), the Natural Science Foundation of Heilongjiang Province (YQ2022F004), the Hong Kong RGC Early Career Scheme (9048212), and the CAAI-Huawei MindSpore Open Fund.

\clearpage

\appendix
\section{Network Architecture}
As mentioned in the main manuscript, the proposed VIDUE consists of an exposure-aware feature extractor $g_e$, an intra- and inter-motion analyzer $g_a$, and a video reconstruction network $f$.
%
% In the following, we first present in detail the architecture of $g_e$ for learning exposure-aware feature representation $\bm u$. We then elaborate the architecture of $g_a$, and the video reconstruction network $f$.
\subsection{Exposure-Aware Feature Extractor}
We illustrate the network structure of the exposure-aware feature extractor $g_e$ in Table~\ref{table:detect}, in which  $\mathrm{ResBlock}$ is detailed in Table~\ref{table:Resblock}. 

\subsection{Intra- and Inter-Motion Analyzer}
 We choose to analyze both intra-motion within each video frame and inter-motion between frames. Our motion analyzer $g_a: \mathbb{R}^{(T\times 3)\times H\times W} \mapsto \mathbb{R}^{T\times 1\times H\times W}\times \mathbb{R}^{(S\times T)\times2\times H\times W}$, computes, from an input video sequence, $T$ intra-motion maps and $(S\times T)\times 2$ inter-motion maps of the same size, respectively. We show the network structure for intra-motion analysis in Table~\ref{table:mmap}.
The network structure for inter-motion analysis is similar to RefineNet proposed in \cite{xu2019quadratic}, which is detailed in Table~\ref{table:ref}.

% \begin{table*}[htb] %\footnotesize
% 	\centering
% 	\setlength{\tabcolsep}{10pt}
% 	\begin{tabular}{cll}
% 		\hline
% 		&\emph{Input1} & Feature maps from the previous layer $\bm{z}$;  \\
% 		&\emph{Input2} & Exposure-aware feature representation $\bm{u}$; \\ \hline
% 		&     & \emph{Input2}   \\
% 		&\emph{Layer 1} & $\mathrm{Linear}(256,64); \mathrm{LeakyReLU}; \mathrm{Linear}(64,\emph{D}); \mathrm{Sigmoid};$\\ 
% 		&\       & $\mathrm{Mul}(\emph{Input1}, \emph{Layer1})$; \\ \hline
% 		%		&\emph{Layer 4} & Conv.($32,32,3,1,1$); ReLU; \\ 
% 		%		&\emph{Layer 5} & Conv.($32,32,3,1,1$); ReLU; \\ 
% 		%		&\       & Add(\emph{Layer3}, \emph{Layer5}); ReLU; \\ \hline
% 		%		&\ ......       &  ......   \\ \hline
% 	\end{tabular}
% 	\caption{The architecture of $\mathrm{SE}$. Fully connected layer is the form $\mathrm{Linear}(input\, channel, output\, channel)$. \emph{D} is the channel dimension of previous layer $\bm{z}$. $\mathrm{Mul}(\cdot)$ is the channel-wise multiplication.}
% 	\label{table:SE}
% \end{table*} 

\subsection{Video Reconstruction Network}
The video reconstruction network $f: \mathbb{R}^{T\times 3\times H\times W}\mapsto\mathbb{R}^{(S\times T)\times 3\times H\times W}$ is built upon the exposure-aware representation $\bm u$ and motion-aware representations $\bm m$ and $\bm n$ by progressive exposure-adaptive convolution and motion refinement. The network structure is illustrated in Table~\ref{table:recon}.

\section{More Results on Benchmark Datasets}
 In this section, we provide more visual results of VIDUE in comparison to existing video interpolation and deblurring methods.
 Please refer to the link for more visual results at \href{https://1drv.ms/u/s!AnS8oR69EOSManDDKpM3etmKVOU?e=McfbLN}{\textbf{\emph{https://onedrive.live.com}}}.
%We also retrain or finetune all methods on our setting for fair comparison. 
%
% \begin{figure*}[!t]\footnotesize
% 	\centering
% 	\setlength{\tabcolsep}{1.5pt}
% 	\begin{tabular}{ccccc}
		
% 		\includegraphics[width=0.28\linewidth]{ablation/71_GOPR0385_11_01_33186/show_img}&  %0.28
% 		\includegraphics[width=0.15\linewidth]{ablation/71_GOPR0385_11_01_33186/wo1}&
% 		\includegraphics[width=0.15\linewidth]{ablation/71_GOPR0385_11_01_33186/wbad1}&
% 		\includegraphics[width=0.15\linewidth]{ablation/71_GOPR0385_11_01_33186/Ours1}&
% 		\includegraphics[width=0.15\linewidth]{ablation/71_GOPR0385_11_01_33186/up1}\\
% 		\multicolumn{1}{c}{\scriptsize Blurred Frame} &
% 		\multicolumn{1}{c}{\scriptsize Exposure-Agnostic} &
% 		\multicolumn{1}{c}{\scriptsize Ordinal Regression} &
% 		\multicolumn{1}{c}{\scriptsize \textbf{Contrastive Learning}} &
% 		\multicolumn{1}{c}{\scriptsize Known Exposure Time}  \\
% 	\end{tabular}
% 	\vspace{-1em}
% 	\caption{Visual comparison for using different exposure-aware feature representations on the GoPro dataset, corresponding to Table~4 in main manuscript. The default setting is highlighted in bold.}
% 	\label{fig:tp}
% \end{figure*} %Supplementary
% \begin{figure*}[!t]\footnotesize
% 	\setlength{\tabcolsep}{1.5pt}
% 	\begin{tabular}{cccccc}
% 		\includegraphics[width=1\linewidth]{ablation/ablation_854_11_00_55}\\
% 	\end{tabular}
% 	\vspace{-2em}
% 	\caption{Visual comparison for each component on the GoPro dataset. Each result corresponds to the rows in Table~5 of main manuscript.}
% 	\label{fig:ablation}
% \end{figure*} %Supplementary
\subsection{More Results on the GoPro Dataset}
On the GoPro dataset with mild blur, VIDUE generally achieves more visually appealing results with sharper boundaries and fewer artifacts, as shown in Figs.~\ref{fig:gopro8x_} and~\ref{fig:gopro8x_full}.

%Our VIDUE can benefit from learned temporal conditions, which can condition the video restoration according to exposure time. 
%%
%Different from them, our method can condition video restoration under different exposure time settings according to the learned temporal style, which leads to an optimal result of different settings.
%%
%Above all, 
\subsection{More Results on the Adobe Dataset}
On the Adobe dataset with heavy blur, VIDUE still performs favorably as shown in Figs.~\ref{fig:adobe8x_} and~\ref{fig:adobe8x_full}.

\subsection{More Results on the RealBlur Dataset}
On the RealBlur dataset, VIDUE generates the least artifacts and sharper textures compared with other methods, as shown in Figs.~\ref{fig:real_} and~\ref{fig:real8x_full}. 

\subsection{More Results on Video $\times 16$ 
 Interpolation and Deblurring }
Although video $\times 16$ interpolation and deblurring is extremely difficult (without the help of extra auxiliary information), VIDUE is better at it than others, as shown in Figs.~\ref{fig:gopro16x} and~\ref{fig:gopro16x_full}.
%
%Please note that it is laborious but worth getting results for the SOTA methods in terms of multi-frame interpolation ($\times 8$) and multi-frame interpolation ($\times 16$).

%%%%%%%%% REFERENCES

{\small
\bibliographystyle{ieee_fullname}
% \balance
\bibliography{egbib}
}
% \clearpage

\begin{table*}[htb] %\footnotesize
	\centering
        \renewcommand\thetable{S1}
	\setlength{\tabcolsep}{10pt}
	\begin{tabular}{cll}
		\toprule
		&\emph{Input} & A sequence of blurred frames ($\mathrm{indim} = T\times 3$) \\ \hline
		&\emph{Layer 1} & $\mathrm{Conv}(\mathrm{indim},64,7,1,3); \mathrm{BN}; \mathrm{LeakyReLU}; \mathrm{ResBlock}(64,64,3,1,1,\mathrm{BN=True}) \times 2 $ \\ 
		&\emph{Layer 2} & $\mathrm{Conv}(64,128,3,2,1); \mathrm{BN}; \mathrm{LeakyReLU}; \mathrm{ResBlock}(128,128,3,1,1,\mathrm{BN=True}) \times 2 $ \\ 
		&\emph{Layer 3} & $\mathrm{Conv}(128,256,3,2,1); \mathrm{BN}; \mathrm{LeakyReLU}; \mathrm{ResBlock}(256,256,3,1,1,\mathrm{BN=True}) \times 2 $ \\ 
		&\emph{Layer 4} & $\mathrm{Conv}(256,256,3,2,1); \mathrm{BN}; \mathrm{LeakyReLU}; \mathrm{ResBlock}(256,256,3,1,1,\mathrm{BN=True}) \times 2 $ \\ 
		&\emph{Layer 5}  & $\mathrm{GlobalAvgPool}$ \\ 
		&\emph{Layer 6}  & $\mathrm{Linear}(256,1024); \mathrm{LeakyReLU}; \mathrm{Linear}(1024,256)$ \\ 
		\hline
		%		&\emph{Layer 4} & Conv.($32,32,3,1,1$); ReLU; \\ 
		%		&\emph{Layer 5} & Conv.($32,32,3,1,1$); ReLU; \\ 
		%		&\       & Add(\emph{Layer3}, \emph{Layer5}); ReLU; \\
		&\emph{Output} & Exposure-aware feature representation $\bm{u}$ \\
		\bottomrule
		%		&\ ......       &  ......   \\ \hline
	\end{tabular}
	\caption{The structure of the exposure-aware feature extractor $g_e$. $T$ is the length of the input video sequence. $\mathrm{BN}$ stands for batch normalization. Convolution is in the form of (input\, channel, output\, channel, kernel\, size, stride, padding). The linear (\ie, fully-connected) layer  is in the form of (input\, channel, output\, channel). }
	\label{table:detect}
\end{table*}
\begin{table*}[htb] %\footnotesize
	\centering
        \renewcommand\thetable{S2}
	\setlength{\tabcolsep}{10pt}
	\begin{tabular}{cll}
		\toprule
		% &\emph{Input} & Feature maps from the previous layer \\ \hline
		&\emph{Layer 1} & $\mathrm{Conv}(D,D,3,1,1); \mathrm{BN}; \mathrm{LeakyReLU}$ \\ 
		&\emph{Layer 2} & $\mathrm{Conv}(D,D,3,1,1); \mathrm{BN}$  \\ 
		&\       & Residual connection between the input and the output of \emph{Layer 2} via addition \\ \bottomrule
		%		&\emph{Layer 4} & Conv.($32,32,3,1,1$); ReLU; \\ 
		%		&\emph{Layer 5} & Conv.($32,32,3,1,1$); ReLU; \\ 
		%		&\       & Add(\emph{Layer3}, \emph{Layer5}); ReLU; \\ \hline
		%		&\ ......       &  ......   \\ \hline
	\end{tabular}
	\caption{The structure of $\mathrm{ResBlock}$. We may remove  $\mathrm{BN}$ by setting the corresponding indicator variable to $\mathrm{False}$.
		%		 $\mathrm{ResBlock}(\cdots,\mathrm{BN=False})$. 
		$D$ is the channel dimension of the feature maps from the previous layer. }
	\label{table:Resblock}
\end{table*} 

\begin{table*}[htb] %\footnotesize
	\centering
        \renewcommand\thetable{S3}
	\setlength{\tabcolsep}{10pt}
	\begin{tabular}{cll}
		\toprule
		&\emph{Input} & One blurred frame ($\mathrm{indim} = 3$) \\ \hline
		&\emph{Layer 1} & $\mathrm{PixelUnshuffle}(2)$ \\
		&\emph{Layer 2} & $\mathrm{Conv}(\mathrm{indim}\times 4,16,5,1,2); \mathrm{ResBlock}(16,16,5,1,2,\mathrm{BN}=\mathrm{False})\times 3$ \\ 
		&\emph{Layer 3} & $\mathrm{Conv}(16,32,5,2,(1,2,1,2)); \mathrm{ResBlock}(32,32,5,1,2,\mathrm{BN}=\mathrm{False})\times 3$ \\ 
		&\emph{Layer 4} & $\mathrm{Conv}(32,64,5,2,(1,2,1,2)); \mathrm{ResBlock}(64,64,5,1,2,\mathrm{BN}=\mathrm{False})\times 3$ \\  
		&\emph{Layer 5} & $\mathrm{Conv}(64,128,1,1,0); \mathrm{LeakyReLU}; \mathrm{Conv}(128,64,3,1,1); \mathrm{LeakyReLU}$ \\ 
		&\emph{Layer 6} & $\mathrm{SE}(256,64,4); \mathrm{TransConv}(64,32,4,2,1); \mathrm{LeakyReLU}$\\
		&    & Concatenation of outputs from \emph{Layer 6} and \emph{Layer 3} \\
		&\emph{Layer 7}  & $\mathrm{Conv}(64,128,1,1,0); \mathrm{LeakyReLU}; \mathrm{Conv}(128,64,3,1,1); \mathrm{LeakyReLU}$ \\ 
		&\emph{Layer 8}  & $\mathrm{SE}(256,64,4); \mathrm{TransConv}(64,16,4,2,1); \mathrm{LeakyReLU}$\\ 
		&    & Concatenation of outputs from \emph{Layer 8} and \emph{Layer 2} \\
		&\emph{Layer 9}  & $\mathrm{Conv}(32,64,1,1,0); \mathrm{LeakyReLU}; \mathrm{Conv}(64,32,3,1,1); \mathrm{LeakyReLU}$ \\ 
		&\emph{Layer 10}  & $\mathrm{SE}(256,32,4); \mathrm{TransConv}(32,32,4,2,1); \mathrm{LeakyReLU}$ \\ 
		&\emph{Layer 11}  & $\mathrm{Conv}(32,4,5,1,2)$ \\ 
		\hline
		&\emph{Output} & Motion offsets \\
		\bottomrule
		%		&\ ......       &  ......   \\ \hline
	\end{tabular}
	\caption{The structure of intra-motion analysis. Adapting to $\bm{u}$ via gain tuning is in the form of squeeze-and-excitation ($\mathrm{SE}$) (vec\, channel, input\, channel, reduction). Transposed convolution is in the same format as convolution.}
	\label{table:mmap}
\end{table*}

\begin{table*}[htb] %\footnotesize
	\centering
        \renewcommand\thetable{S4}
	\setlength{\tabcolsep}{10pt}
	\begin{tabular}{cll}
		\toprule
		&\emph{Input} & A sequence of motion offsets ($\mathrm{indim} = T\times 4$) \\ \hline
		&\emph{Layer 1} & $\mathrm{Conv}(\mathrm{indim},32,7,1,3); \mathrm{LeakyReLU}; \mathrm{Conv}(32,32,7,1,3); \mathrm{LeakyReLU}$ \\ 
		&\emph{Layer 2} & $\mathrm{AvgPool}(2); \mathrm{Conv}(32,64,5,1,2); \mathrm{LeakyReLU}; \mathrm{Conv}(64,64,5,1,2); \mathrm{LeakyReLU}$ \\ 
		&\emph{Layer 3} & $\mathrm{AvgPool}(2); \mathrm{Conv}(64,128,3,1,1); \mathrm{LeakyReLU}; \mathrm{Conv}(128,128,3,1,1); \mathrm{LeakyReLU} $\\ 
		&\emph{Layer 4} & $\mathrm{AvgPool}(2); \mathrm{Conv}(128,256,3,1,1); \mathrm{LeakyReLU}; \mathrm{Conv}(256,256,3,1,1); \mathrm{LeakyReLU}$ \\   
		&\emph{Layer 5} & $\mathrm{AvgPool}(2); \mathrm{Conv}(256,512,3,1,1); \mathrm{LeakyReLU}; \mathrm{Conv}(512,512,3,1,1); \mathrm{LeakyReLU}$\\
		&\emph{Layer 6} & $\mathrm{AvgPool}(2); \mathrm{Conv}(512,512,3,1,1); \mathrm{LeakyReLU}; \mathrm{Conv}(512,512,3,1,1); \mathrm{LeakyReLU}$\\
		&\emph{Layer 7}  & $\mathrm{Up}(2); \mathrm{Conv}(512,512,3,1,1); \mathrm{LeakyReLU}; \mathrm{Conv}(512,512,3,1,1); \mathrm{LeakyReLU}; \mathrm{SE}(256,512,4)$\\ 
		&\emph{Layer 8}  & $\mathrm{Up}(2); \mathrm{Conv}(512,256,3,1,1); \mathrm{LeakyReLU}; \mathrm{Conv}(256,256,3,1,1); \mathrm{LeakyReLU}; \mathrm{SE}(256,256,4)$\\ 
		&\emph{Layer 9}  & $\mathrm{Up}(2); \mathrm{Conv}(256,128,3,1,1); \mathrm{LeakyReLU}; \mathrm{Conv}(256,128,3,1,1); \mathrm{LeakyReLU}; \mathrm{SE}(256,128,4)$\\ 
		&\emph{Layer 10}  & $\mathrm{Up}(2); \mathrm{Conv}(128,64,3,1,1); \mathrm{LeakyReLU}; \mathrm{Conv}(64,64,3,1,1); \mathrm{LeakyReLU}; \mathrm{SE}(256,64,4)$\\
		&\emph{Layer 11}  & $\mathrm{Up}(2);  \mathrm{Conv}(64,32,3,1,1); \mathrm{LeakyReLU};  \mathrm{Conv}(32,32,3,1,1); \mathrm{LeakyReLU}; \mathrm{SE}(256,32,4)$\\
		&\emph{Layer 12}  &$\mathrm{Conv}(32,S\times T\times 2,3,1,1)$ \\
		\hline
		&\emph{Output} & Inter-motion map $\bm{n}$ \\
		\bottomrule
	\end{tabular}
	\caption{The structure of inter-motion analysis. $\mathrm{Up}(\cdot)$ stands for bilinear interpolation.}
	\label{table:ref}
\end{table*}

\begin{table*}[htb] %\footnotesize
	\centering
        \renewcommand\thetable{S5}
	\setlength{\tabcolsep}{10pt}
	\begin{tabular}{cll}
		\toprule
		&\emph{Input} & A sequence of blurred frames ($\mathrm{indim} = T\times 3$) \\ \hline
		&\emph{Layer 1} & $\mathrm{Conv}(\mathrm{indim},64,7,1,3); \mathrm{LeakyReLU}; \mathrm{ResBlock}(64,64,3,1,1,\mathrm{BN}=\mathrm{False})\times 3$  \\ 
		&\emph{Layer 2} & $\mathrm{Conv}(64,128,3,2,1); \mathrm{LeakyReLU}; \mathrm{ResBlock}(128,128,3,1,1,\mathrm{BN}=\mathrm{False})\times 6$ \\ 
		&\emph{Layer 3} & $\mathrm{Conv}(128,256,3,2,1); \mathrm{LeakyReLU}; \mathrm{ResBlock}(256,256,3,1,1,\mathrm{BN}=\mathrm{False})\times 6$ \\ 
		&\emph{Layer 4} & $\mathrm{Conv}(256,256,3,2,1); \mathrm{LeakyReLU}; \mathrm{ResBlock}(256,256,3,1,1,\mathrm{BN}=\mathrm{False})\times 6$ \\ 
		&\emph{Layer 5}  & $\mathrm{TransConv}(256,256,3,2,1); \mathrm{LeakyReLU}; \mathrm{ResBlock}(256,256,3,1,1,\mathrm{BN}=\mathrm{False})\times 6$ \\ 
		&\emph{Layer 6}  & $\mathrm{E\_Conv}(256,256,4)$ \\
		&\emph{Layer 7} & $\mathrm{Motion\_Refine}(256,3,S\times T); \mathrm{Up}(2)$ \\
		&  & Addition of outputs from \emph{Layer 6} and \emph{Layer 3} \\ 
		&\emph{Layer 8}& $\mathrm{TransConv}(256,128,3,2,1); \mathrm{LeakyReLU}; \mathrm{ResBlock}(128,128,3,1,1,\mathrm{BN}=\mathrm{False})\times 6$ \\ 
		&\emph{Layer 9}  & $\mathrm{E\_Conv}(256,128,4)$ \\
		&\emph{Layer 10} & $\mathrm{Motion\_Refine}(128,3,S\times T)$ \\
		&  & Addition of outputs from \emph{Layer 10} and \emph{Layer 7} \\ 
            &\emph{Layer 11}  & $\mathrm{Up}(2)$  \\
		&    & Addition of outputs from \emph{Layer 9} and \emph{Layer 2} \\
		&\emph{Layer 12}  & $\mathrm{TransConv}(128,64,3,2,1); \mathrm{LeakyReLU}; \mathrm{ResBlock}(64,64,3,1,1,\mathrm{BN}=\mathrm{False})\times 6$ \\
		&\emph{Layer 13}  & $\mathrm{E\_Conv}(256,64,4)$ \\
		&\emph{Layer 14} & $\mathrm{Motion\_Refine}(64,3,S\times T)$ \\
		&\emph{Layer 15}  & Addition of outputs from \emph{Layer 14} and \emph{Layer 11} \\
		&  & Addition of outputs from \emph{Layer 13} and \emph{Layer 1} \\ 
		&\emph{Layer 16}  & $\mathrm{ResBlock}(64,64,3,1,1,\mathrm{BN}=\mathrm{False})\times 3; \mathrm{Conv}(64,S\times T\times 3,3,1,1); \mathrm{Split}(3)$ \\
		&\emph{Layer 17}  & Addition of outputs from \emph{Layer 16} and \emph{Layer 15} \\
		\hline
		%		&\emph{Layer 4} & Conv.($32,32,3,1,1$); ReLU; \\ 
		%		&\emph{Layer 5} & Conv.($32,32,3,1,1$); ReLU; \\ 
		%		&\       & Add(\emph{Layer3}, \emph{Layer5}); ReLU; \\
		&\emph{Output} & Final sharp video sequence $\hat{\bm{x}}$ \\
		\bottomrule
		%		&\ ......       &  ......   \\ \hline
	\end{tabular}
	\caption{The architecture of video reconstruction network. Adapting to $\bm{u}$ via exposure-aware convolution is in the form $\mathrm{E\_Conv}$ (vec\, channel, input\, channel, reduction). Motion refinement is in the form of $\mathrm{Motion\_Refine}$ (input\, channel, kernel\, size, output\, sequence). $\mathrm{Split}(\cdot)$ stands for the channel splitting operation.}
	\label{table:recon}
\end{table*}

\begin{figure*}[!t]\footnotesize
        \renewcommand\thefigure{S1}
	\setlength{\tabcolsep}{1.5pt}
	\begin{tabular}{cccccc}
		\includegraphics[width=1\linewidth]{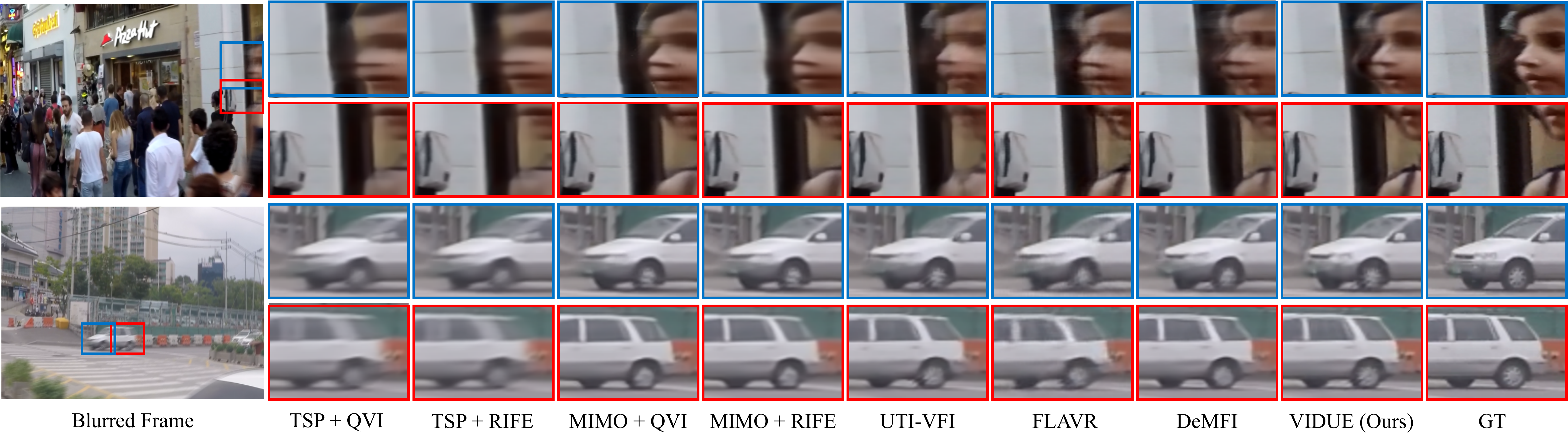}\\
	\end{tabular}
	\vspace{-2em}
	\caption{More visual results on GoPro \cite{Nah_2017_CVPR} ($\times 8$). Blue and red patches are from deblurred and interpolated frames, respectively. TSP is short for CDVD-TSP, and MIMO is short for MIMOUNetPlus.}
	\label{fig:gopro8x_}
\end{figure*} %Supplementary

\begin{figure*}[!t]\footnotesize
        \renewcommand\thefigure{S2}
	\setlength{\tabcolsep}{1.5pt}
	\begin{tabular}{cccccc}
		\includegraphics[width=1\linewidth]{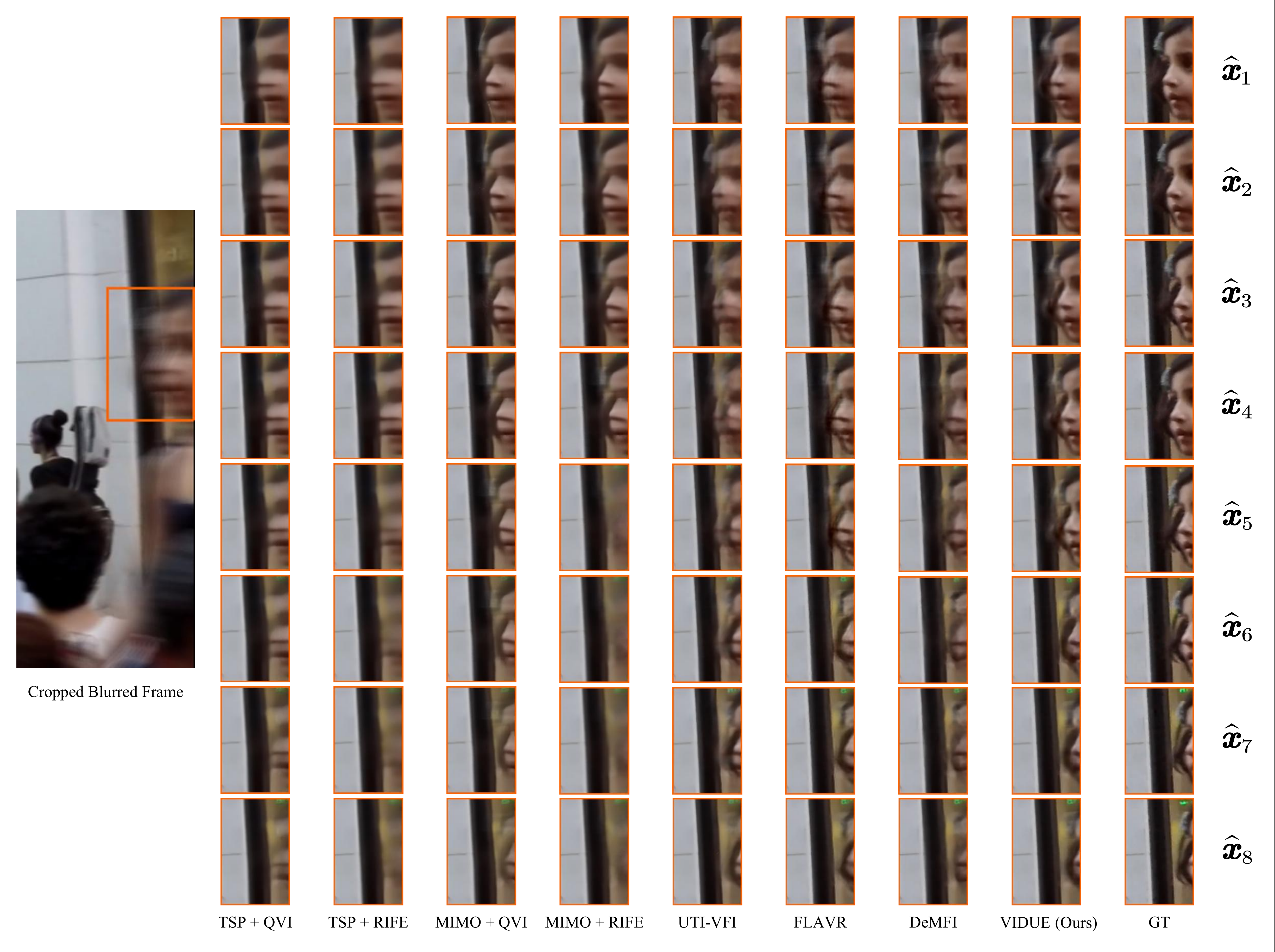}\\
	\end{tabular}
	\vspace{-2em}
	\caption{More visual results on GoPro \cite{Nah_2017_CVPR} ($\times 8$).}
	\label{fig:gopro8x_full}
\end{figure*} %Supplementary

\begin{figure*}[!t]\footnotesize
        \renewcommand\thefigure{S3}
	\setlength{\tabcolsep}{1.5pt}
	\begin{tabular}{cccccc}
		\includegraphics[width=1\linewidth]{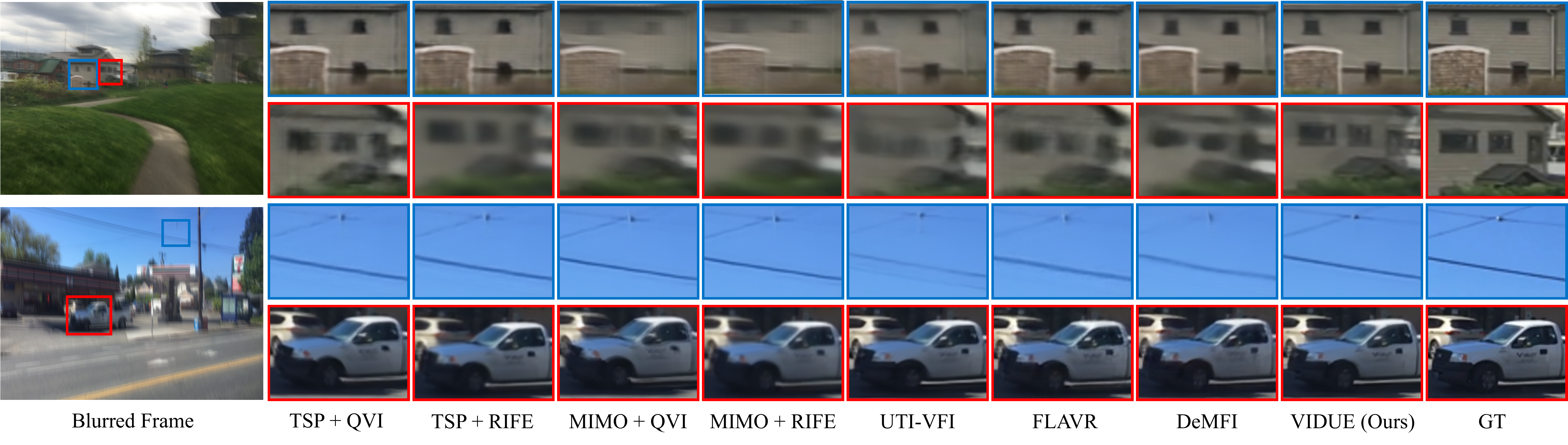}\\
	\end{tabular}
	\vspace{-2em}
	\caption{More visual results on Adobe \cite{su2017deep} ($\times 8$). Blue and red patches are from deblurred and interpolated frames, respectively.}
	\label{fig:adobe8x_}
\end{figure*} %Supplementary

\begin{figure*}[!t]\footnotesize
        \renewcommand\thefigure{S4}
	\setlength{\tabcolsep}{1.5pt}
	\begin{tabular}{cccccc}
		\includegraphics[width=1\linewidth]{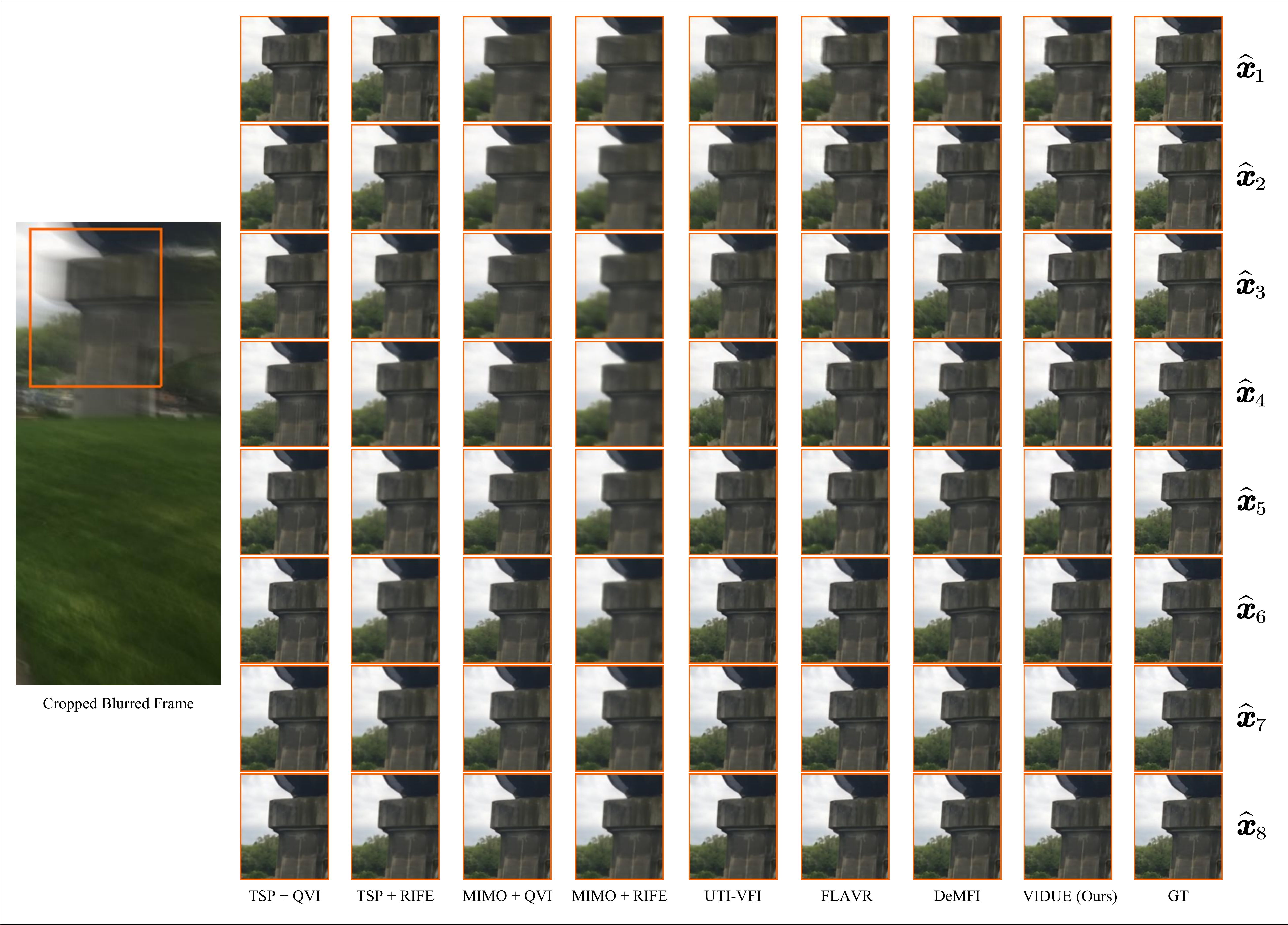}\\
	\end{tabular}
	\vspace{-2em}
	\caption{More visual results on Adobe \cite{su2017deep} ($\times 8$).}
	\label{fig:adobe8x_full}
\end{figure*} %Supplementary

\begin{figure*}[!t]\footnotesize
        \renewcommand\thefigure{S5}
	\setlength{\tabcolsep}{1.5pt}
	\begin{tabular}{cccccc}
		\includegraphics[width=1\linewidth]{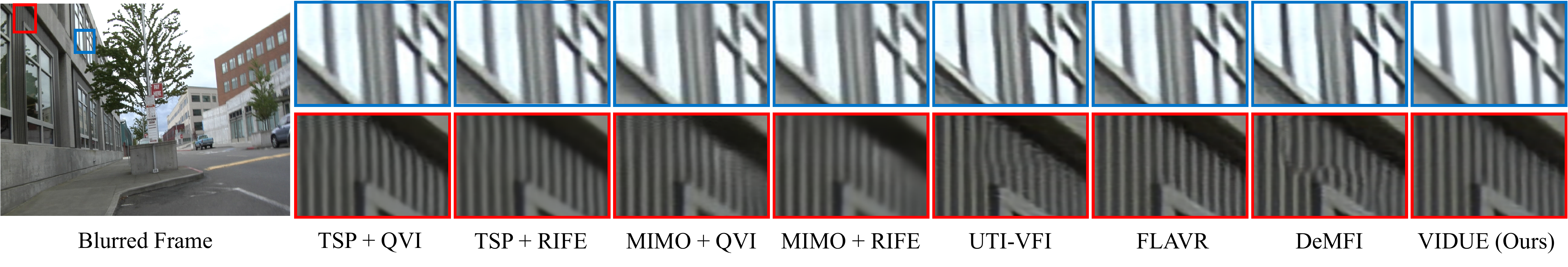}\\
	\end{tabular}
	\vspace{-2em}
	\caption{More visual results on RealBlur \cite{su2017deep} ($\times 8$). Blue and red patches are from deblurred and interpolated frames, respectively.}
	\label{fig:real_}
\end{figure*} %Supplementary
\begin{figure*}[!t]\footnotesize
        \renewcommand\thefigure{S6}
	\setlength{\tabcolsep}{1.5pt}
	\begin{tabular}{cccccc}
		\includegraphics[width=1\linewidth]{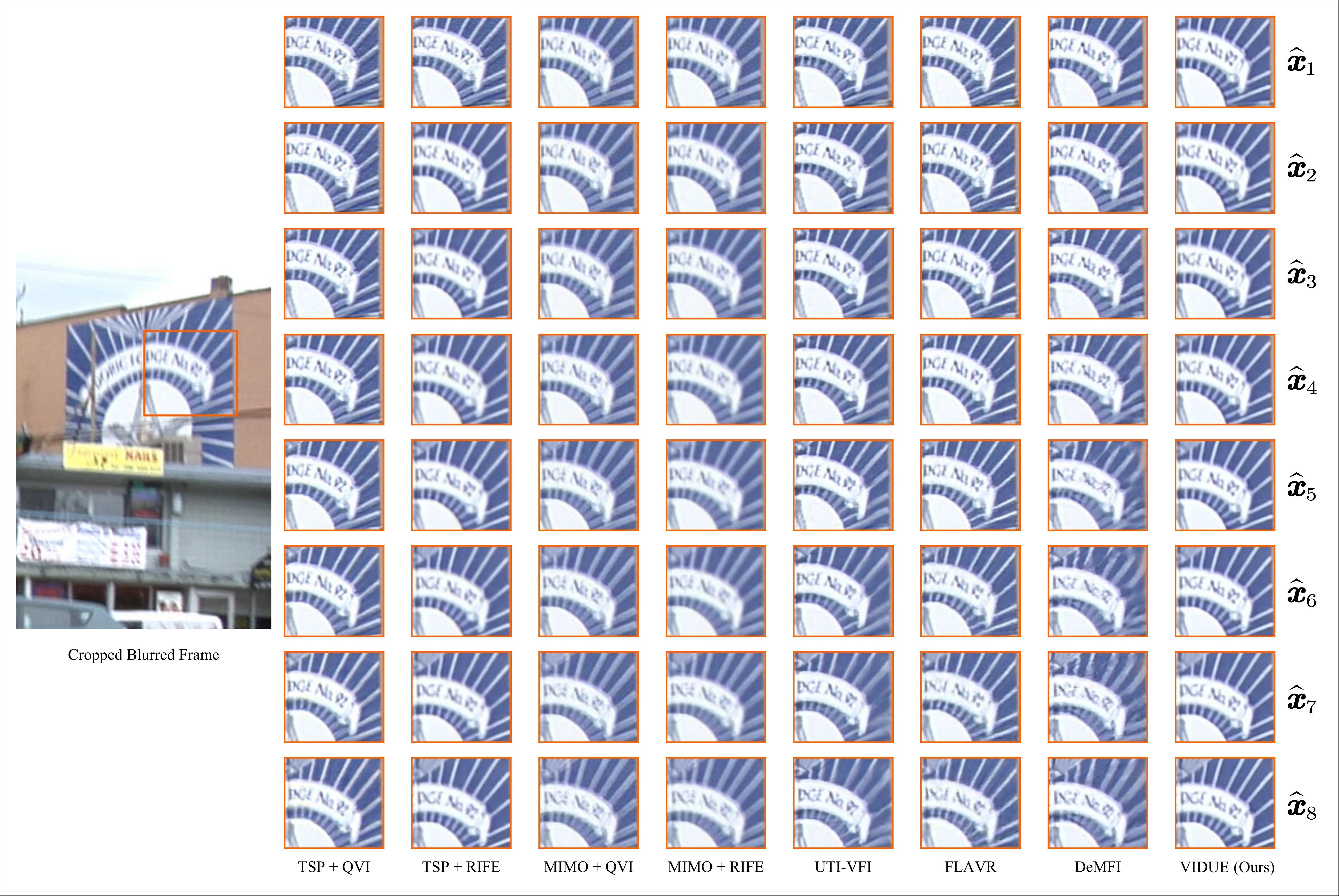}\\
	\end{tabular}
	\vspace{-2em}
	\caption{More visual results on RealBlur \cite{su2017deep} ($\times 8$). }
	\label{fig:real8x_full}
\end{figure*} %Supplementary

\begin{figure*}[!t]\footnotesize
        \renewcommand\thefigure{S7}
	\setlength{\tabcolsep}{1.5pt}
	\begin{tabular}{cccccc}
		\includegraphics[width=1\linewidth]{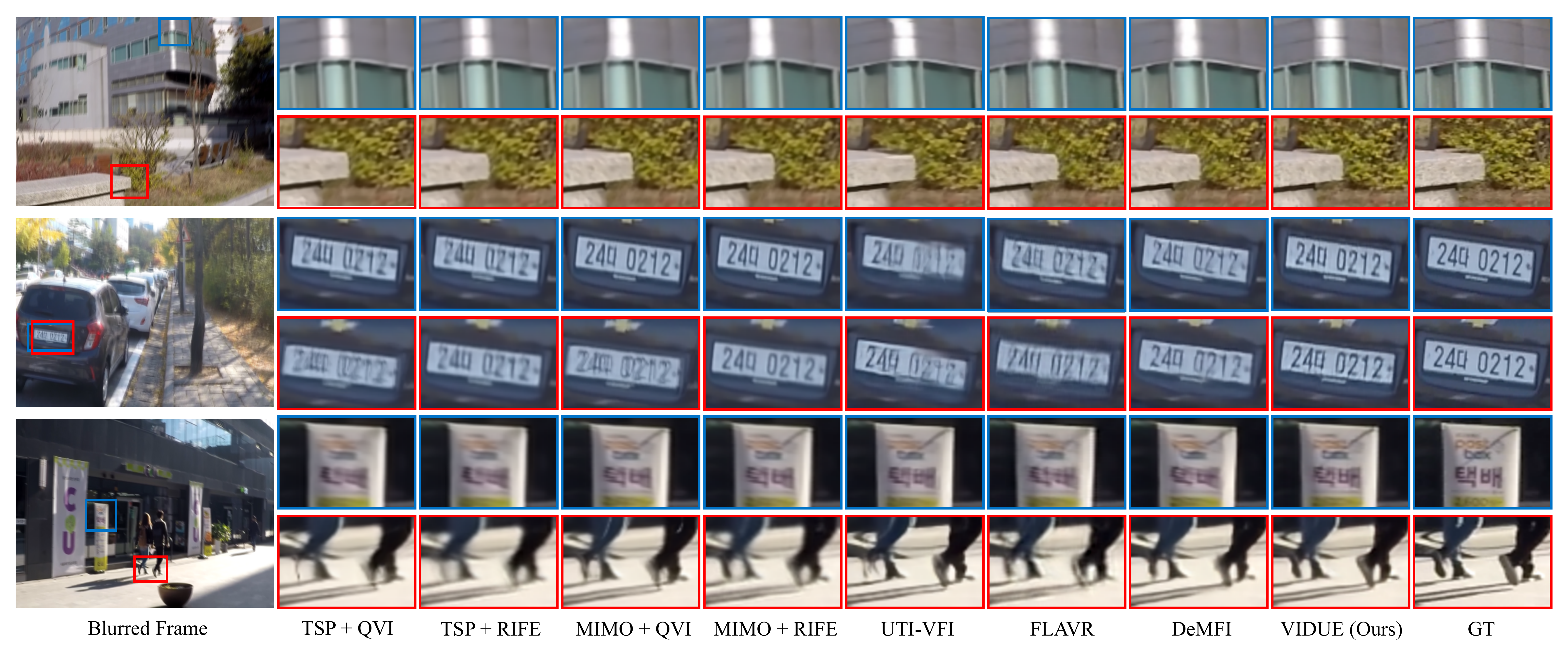}\\
	\end{tabular}
	\vspace{-2em}
	\caption{More visual results on GoPro \cite{Nah_2017_CVPR} ($\times 16$). Blue and red patches are from deblurred and interpolated frames, respectively.}
	\label{fig:gopro16x}
\end{figure*} %Supplementary
\begin{figure*}[!t]\footnotesize
        \renewcommand\thefigure{S8}
	\setlength{\tabcolsep}{1.5pt}
	\begin{tabular}{cccccc}
		\includegraphics[width=1\linewidth]{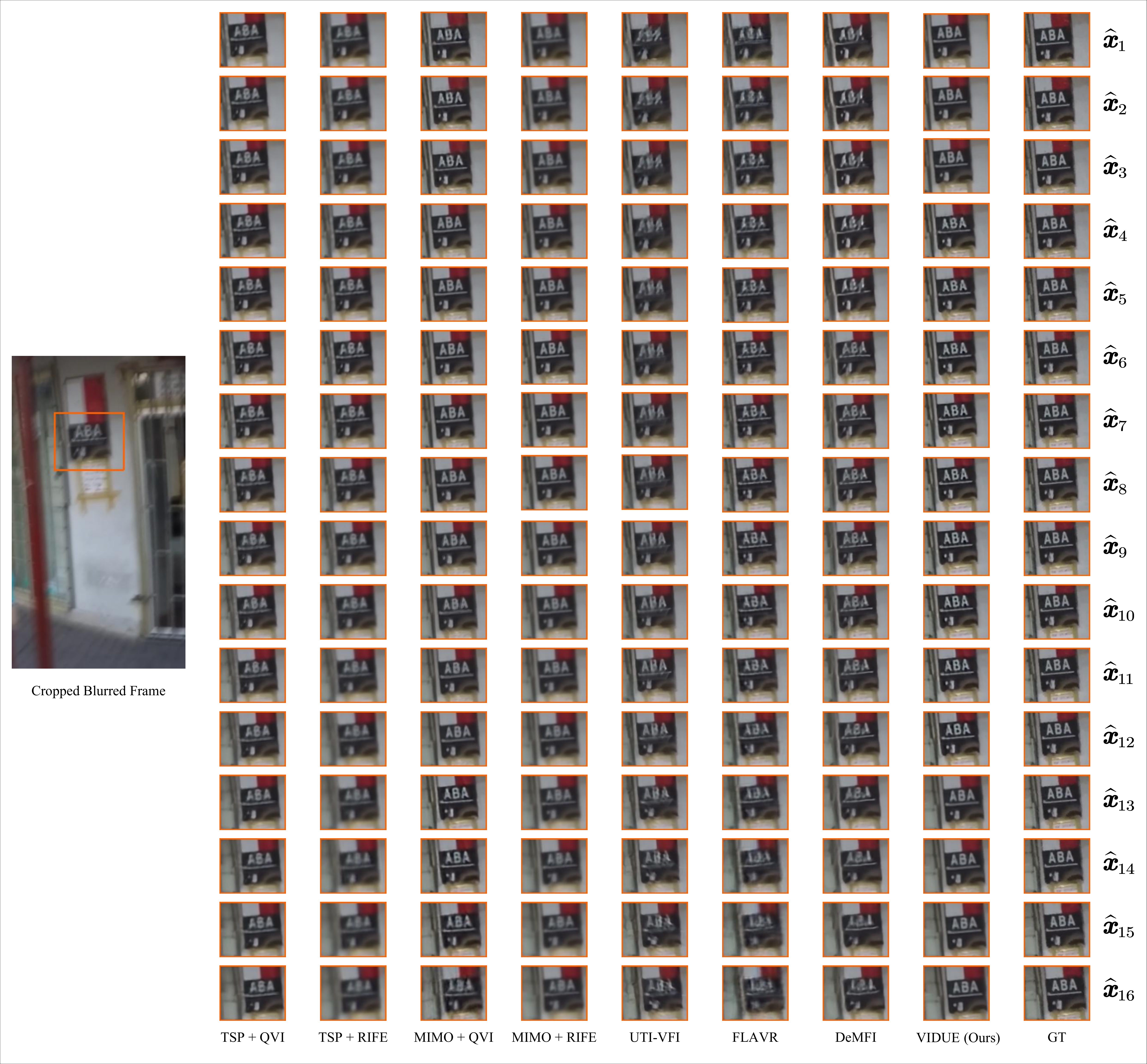}\\
	\end{tabular}
	\vspace{-2em}
	\caption{More visual results on GoPro \cite{Nah_2017_CVPR} ($\times 16$).}
	\label{fig:gopro16x_full}
\end{figure*} %Supplementary

\end{document}